\documentclass{article} 
\usepackage{iclr2027_conference,times}

\usepackage{amsmath,amsfonts,bm}

\def\eqref#1{equation~\ref{#1}}

\def\1{\bm{1}}

\DeclareMathAlphabet{\mathsfit}{\encodingdefault}{\sfdefault}{m}{sl}
\SetMathAlphabet{\mathsfit}{bold}{\encodingdefault}{\sfdefault}{bx}{n}

\usepackage{hyperref}
\usepackage{url}

\usepackage{microtype}
\usepackage{amsmath,amssymb,mathtools}
\usepackage{booktabs}
\usepackage{graphicx}
\usepackage{wrapfig}
\usepackage[table]{xcolor}
\usepackage{makecell}
\usepackage{multirow}
\usepackage{pifont}

\definecolor{roboflblue}{HTML}{1F4E79}
\definecolor{roboflblueLight}{HTML}{EAF2F8}
\definecolor{roboflgreen}{HTML}{E8F5E9}
\definecolor{roboflgray}{HTML}{F4F6F7}

\title{RoboFL: Federated Expert Assembly for World Action Models}

\author{Rongyu Zhang\thanks{Equal contributions. $^\dagger$ Corresponding authors. $^\ddagger$ State Key Laboratory of Multimedia Information Processing, School of Computer Science, Peking University} \\
Nanjing University\\
\And
Ruizhi Fan$^\ast$ \\
Nanjing University\\
\And
Yunfan Lou \\
Peking University \\
\And
Hengyu Fang \\
Nanjing University \\
\And
Shenli Zhang \\
Nanjing University \\
\And
Chenrui Wu \\
Simon Fraser University \\
\And
Yili Jin \\
Simon Fraser University \\
\And
Li Du \\
Nanjing University \\
\And
Dan Wang \\
Hong Kong University of Science and Technology \\
\And
Yuan Du$^\dagger$ \\
Nanjing University \\
\And
Shanghang Zhang$^\dagger$ \\
Peking University$^\ddagger$ \\
}

\newcommand{\robofl}{\textsc{RoboFL}}
\newcommand{\mosaic}{\textsc{MoSAIC}}
\newcommand{\fard}{\textsc{FARD}}
\newcommand{\pcea}{\textsc{PCEA}}
\newcommand{\taskname}[1]{\rotatebox[origin=c]{60}{\scriptsize\texttt{#1}}}
\iclrfinalcopy 
\begin{document}

\maketitle

\begin{figure*}[h]
  \centering
  \includegraphics[width=\textwidth]{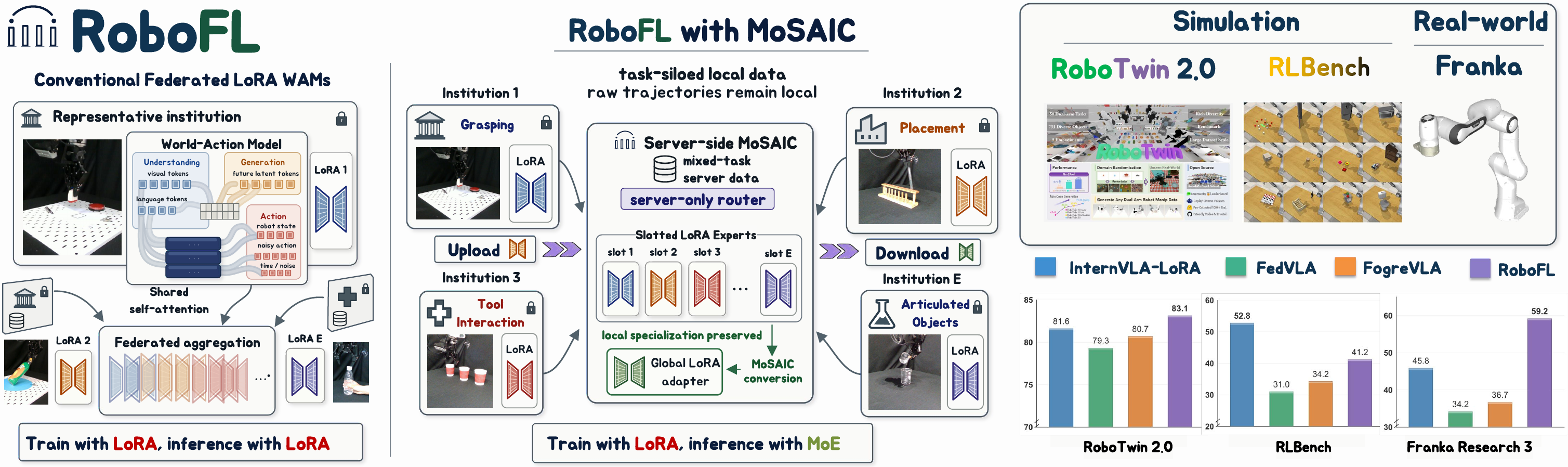}
  \caption{
    \textbf{From federated parameter averaging to prior-informed expert assembly.}
    \textbf{Left:} Aggregating independently trained LoRA-MoE models can dilute local specialization and introduce inconsistent routing across heterogeneous clients.
    \textbf{Middle:} \robofl \ organizes institutions into task silos and directly installs their task-trained LoRA into the expert branches of a server MoE.
    \textbf{Right:} \robofl \ is evaluated in the RoboTwin 2.0 and RLBench simulator, and on a real-world Franka robot.
}
  \label{fig:teaser}
\end{figure*}

\begin{abstract}
Vision-language-action and world-action models are increasingly popular, yet remain bottlenecked by physical interaction data that is scarce, institutionally siloed, and task-heterogeneous. A natural federated solution is to let each client adapt a shared foundation model through parameter-efficient fine-tuning, avoiding the exchange of full-model updates. However, federating these adapters is nontrivial, as naive aggregation can entangle incompatible updates, while incorporating MoE-style routing into federated aggregation may dilute specialization and destabilize expert selection.
We present \robofl, which instantiates \mosaic \ (Mixture of Slotted Adapters) for federated world-action learning. MoSAIC directly installs locally trained LoRA adapters as the expert branches of a server MoE. Server-side routers learn token assignments over these prior-informed branches while jointly refining routing and expert parameters. Foresight-to-Action Routing Distillation (\fard) aligns routing across the model's three paths, while Path-Consensus Expert Aggregation (\pcea) converts complete expert updates into a compact global adapter for personalized redistribution. Experiments on RoboTwin 2.0, RLBench, and a real-world Franka robot arm show the superiority of \robofl \ with structured expert assembly, as it outperforms centralized PEFT InternVLA-A1 by 12.23\% on the Franka arm, while reducing per-round client communication by up to 86.81\% relative to MoE-based federated VLA baselines.
\end{abstract}

\section{Introduction}

Vision-language-action (VLA) models~\citep{brohan2023rt,driess2023palme,octomodelteam2024octo,kim2025openvla,black2025pi0} and world-action models (WAMs)~\citep{cai2026internvla,bi2026motus,ye2026wam} unify perception, language understanding, future prediction, and control in a single policy. Their progress has been driven by broader multimodal and multi-robot training, flow-based action generation, and future-state modeling. However, general-purpose embodied intelligence remains constrained by real interaction data. Unlike web data, robot demonstrations require hardware, expert teleoperation, repeated resets, safety monitoring, and synchronized multimodal sensing. They are expensive to collect, especially for contact-rich, long-horizon, and multi-embodiment tasks, making high-quality trajectories both scarce and valuable~\citep{wu2025robomind,luo2025robobench}.

This data bottleneck is compounded by privacy and ownership constraints. Robot trajectories can expose private environments, user routines, and proprietary procedures, and their substantial acquisition cost makes them valuable institutional assets that owners are reluctant to share. Federated learning (FL)~\citep{mcmahan2017communication,kairouz2021advances} therefore provides a natural infrastructure for large-scale embodied training, letting institutions contribute complementary experience without centralizing raw data. This task-silo setting is practical rather than artificial: an institution typically owns the hardware, expertise, and suite of tasks for a focused family of manipulation tasks, so disjoint silos arise directly from existing practice. The resulting federation is nevertheless highly heterogeneous: client task families are largely disjoint and uneven in size, so each client optimizes over a different label distribution, and the aggregate problem is strongly non-IID. Under such task-level skew, local optima drift apart, and naive averaging blends transformations that are valid only within their own families. \textbf{The objective is thus to combine complementary task-specific capabilities into a broadly competent policy without exposing or erasing their source.}

Mixture-of-Experts (MoE)~\citep{jacobs1991adaptive,shazeer2017sparsely,fedus2022switch} provides a natural representation for such task diversity, combining specialized experts with a shared backbone and input-dependent routing. For example, FedVLA~\citep{miao2025fedvla} trains dual-gating MoE models on clients and aggregates their aligned-MoE trunk parameters for performance enhancement. Yet a modular architecture does not, by itself, ensure coherent federated learning. When each client trains its own MoE on a narrow task distribution, the corresponding expert branches can acquire distinct specializations, while the routers learn assignments shaped by local task frequencies. Averaging these expert updates can dilute task-specific knowledge, and aggregating the routers can disrupt the assignments that make that knowledge useful, leading to performance degradation as shown in Section~\ref{sec:experiment}. \textbf{The key is therefore to separate local specialization from the learning of a shared routing policy.}

To address this gap, we propose \robofl, a federated framework that combines a task-silo protocol with server-side expert assembly as shown in Figure~\ref{fig:teaser}. Designed for communication-constrained FL settings, \robofl \ leverages parameter-efficient fine-tuning (PEFT)~\citep{hu2022lora,wu2024mole} to enable clients to exchange compact LoRA adaptations rather than full model updates~\citep{miao2025fedvla,zhou2026forgevla}. Specifically, \ding{202} each institution trains a single LoRA adapter on a focused family of tasks and uploads only its adaptation parameters to the server. \ding{203} The server assigns the uploaded adapters to fixed expert slots and composes them into a layer-wise MoE, where a router is trained to select among task-specialized LoRA experts. \ding{204} After server-side routing and expert refinement, the resulting expert updates are converted into a compact global adapter and blended with each client-associated expert before the next communication round.

This procedure yields a \textbf{Mixture-of-Slotted-Adapters} with integration and conversion (\mosaic). By separating local expert formation from global router learning, \robofl \ enables the router to build on task-trained capabilities rather than on initially unspecialized branches. The fixed client-to-slot mapping preserves parameter correspondence across rounds, while still allowing flexible token-to-expert assignments. This design supports more stable routing without imposing multi-expert training or communication on clients: local training and exchanged adapters retain the footprint of single-adapter fine-tuning, while routing and multi-expert optimization remain on the server.

In addition, contemporary WAMs~\citep{cai2026internvla,bi2026motus,ye2026wam} use aligned mixture-of-transformers (MoT) architectures~\citep{liang2024mixture} whose understanding, visual foresight, and action paths provide complementary evidence about the same interaction. Such a three-path structure of WAMs offers an additional source of guidance for this assembly process. \mosaic \ exploits this structure to inform both expert selection and adapter conversion. Therefore, we propose \textbf{Foresight-to-Action Routing Distillation} (\fard), which distills reliable routing consensus between the understanding and generation paths into the action router, using a detached teacher to guide expert selection for action generation. In addition, \textbf{Path-Consensus Expert Aggregation} (\pcea) uses detached consensus across all three paths to weight complete server-refined LoRA updates and reconstruct a rank-constrained global adapter. Together, these mechanisms use agreement across what the model understands, predicts, and executes to guide which experts it activates during server training and how it consolidates their knowledge for subsequent client training.

Finally, we evaluate \robofl \ on RoboTwin 2.0~\citep{mu2025robotwin}, RLBench~\citep{james2020rlbench}, and a real-world Franka robot arm against centralized PEFT WAMs, as well as FL-based VLA methods, as shown on the right side of Figure~\ref{fig:teaser}. Experiments across simulation and the real world show that \robofl \ outperforms existing federated VLA methods and, in certain scenarios, even centralized PEFT WAMs, as it attains  83.12\% overall on RoboTwin 2.0, 4.2\% above federated averaging and 1.52\% above centralized InternVLA-A1, reaches 59.17\% against 46.94\% for centralized InternVLA on six real-world Franka tasks, and cuts client communication by up to 86.81\% against MoE-based federated VLA baselines.

Our contributions are summarized as follows.
\begin{enumerate}
    \item \textbf{Task-siloed federated expert assembly.} We propose \robofl \ with \mosaic \ to assemble client-trained LoRA adapters into a server-routed MoE, keeping raw trajectories local and retaining single-adapter client training and communication.
    \item \textbf{Foresight-guided action routing.} We introduce \fard \ to distill reliable understanding-generation routing consensus into the action router for MoT-based world action models.
    \item \textbf{Path-consensus adapter conversion.} We introduce \pcea \ to aggregate complete server-refined expert updates using three-path consensus and reconstruct a rank-constrained global adapter for personalized redistribution.
\end{enumerate}

\section{Related Works}

\paragraph{Vision-language-action and world-action models.}
VLA models connect multimodal knowledge to robot control, from RT-2~\citep{brohan2023rt} and PaLM-E~\citep{driess2023palme} to multi-robot policies such as Octo~\citep{octomodelteam2024octo} and OpenVLA~\citep{kim2025openvla}. Subsequent work advances flow-based control~\citep{black2025pi0}, dual-system architectures~\citep{bjorck2025groot}, action tokenization~\citep{pertsch2025fast}, spatial representations~\citep{qu2025spatialvla}, and efficient adaptation~\citep{zhang2026mole}. Cosmos~\citep{agarwal2025cosmos} and recent WAMs~\citep{ye2026gigaworld,li2026omega} further exploit future-state prediction. These systems generally assume centralized training rather than assembling future-aware policies from institutionally distributed task data.

\paragraph{VLA and WAMs with MoE and FL.}
MoE models~\citep{fedus2022switch,zhangmoant} support specialized computation, including expert selection and weighting decoupled in AdaMoE~\citep{shen2025adamoe}, force-aware routing in ForceVLA~\citep{yu2025forcevla}, and layer-wise activation in MoLe-VLA~\citep{zhang2026mole}. FLAME~\citep{betran2025flame} benchmarks robotic FL, while FedVLA~\citep{miao2025fedvla} combines federated training with dual-gating MoE and ForgeVLA~\citep{zhou2026forgevla} addresses unannotated vision-action logs through instruction recovery, contrastive planning, and adaptive aggregation. Training MoE routers on narrow client distributions can nevertheless produce inconsistent expert assignments. \robofl \ separates local adapter specialization from server-side routing and jointly refines uploaded experts and module-specific routers on server data.

\paragraph{Federated PEFT and MoE.}
LoRA~\citep{hu2022lora} provides compact low-rank updates for federated adaptation. FLoRA~\citep{wang2024flora}, FRLoRA~\citep{yan2025frlora}, FedSA-LoRA~\citep{guo2025fedsalora}, and LoRA-A2~\citep{koo2025loraa2} study low-rank aggregation and sharing, while FedFisher~\citep{jhunjhunwala2024fedfisher} addresses one-shot aggregation. Mixture of LoRA Experts routes among adapters~\citep{wu2024mole}, FedMoE~\citep{mei2024fedmoe} integrates client-specific sub-MoEs, and pFedMoAP~\citep{luo2025pfedmoap} exchanges prompt experts for personalization. In contrast, \robofl \ installs single-adapter client updates as server MoE experts without federating selection routers, then uses cross-path consensus to convert complete expert updates into compact adapters for personalized redistribution.

\begin{figure*}[t]
  \centering
  \includegraphics[width=\textwidth]{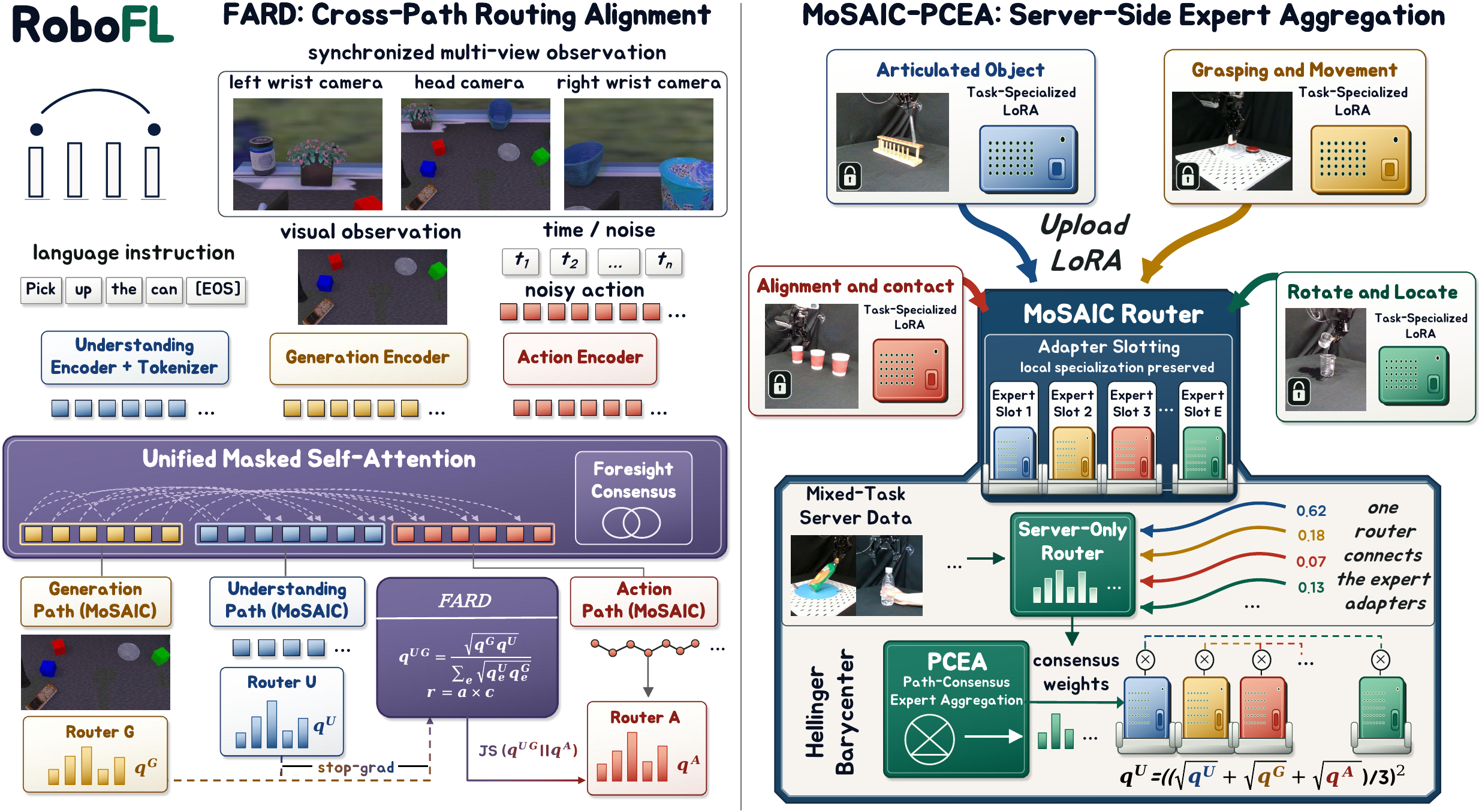}
  \caption{
    \textbf{Overview of \robofl.}
    \textbf{Left:} Modality-specific tokens interact through unified masked self-attention, while \fard \ distills detached understanding-generation consensus into action routing.
    \textbf{Right:} \mosaic \ installs task-trained LoRA adapters as server MoE experts and jointly refines routers and experts.
    \pcea \ aggregates complete expert updates using three-path consensus to form a rank-constrained global adapter, which is blended with each server-refined client expert.
}
  \label{fig:framework}
\end{figure*}

\section{Method}

\subsection{Preliminary}
\paragraph{Three-Path World-Action Architecture.}
\robofl \ is built on a three-path world-action architecture, in which the understanding ($U$), generation ($G$), and action ($A$) paths are coupled through unified masked self-attention. $U$ encodes current images and instructions with Qwen3-VL. $G$ processes Cosmos-tokenized past and current frames to predict future-frame latents. $A$ receives a projected state token, followed by noisy action tokens fused with a time embedding. At each aligned layer, path-specific queries, keys, and values are concatenated along the token dimension for joint attention, then split for path-specific output projections and MLPs. Thus, the paths share the attention computation, not projection or MLP weights. The block-causal mask orders tokens as $U\mid G\mid\text{state}\mid\text{actions}$: each block attends to itself and preceding blocks, subject to padding masks. Action tokens attend bidirectionally within a chunk and use $U/G$ hidden context, not decoded future images.

At matching layer/projection $m$, let $q^P_{m,b}\in\mathbb{R}^{E}$ be the sample-wise mean of the full \emph{pre-top-$k$} router probabilities for path $P\in\{U,G,A\}$. Specifically, pooling masks out invalid observations and language positions in the understanding path $U$ and invalid visual positions in the generation path $G$. In $A$, it includes all action positions, excluding the state token but not action padding. Cross-path comparisons use samples with valid pooled distributions in all three paths.

\paragraph{Flow-Matching Action Policy.}
For a demonstrated action chunk $a\in\mathbb{R}^{H\times d_a}$, noise $\epsilon\sim\mathcal{N}(0,I)$, and time $t\sim p_{\mathrm{time}}$, the action path predicts the conditional velocity
\begin{equation}
\begin{aligned}
    x_t&=(1-t)a+t\epsilon,
    \qquad
    u_t=\epsilon-a,\\
    \mathcal{L}_{\mathrm{action}}
    &=\mathbb{E}_{a,\epsilon,t}
    \left[\operatorname{MSE}_{\mathrm{valid}}\!\left(v_\theta(x_t,t,c),u_t\right)\right],
\end{aligned}
\end{equation}
where $c$ denotes $U/G$ context and the robot state, and MSE averages over actual action dimensions and non-padded steps when padding labels are available. At inference, explicit Euler integration proceeds from $x_1\sim\mathcal{N}(0,I)$ at $t=1$ to $t=0$. The generation loss $\mathcal{L}_{\mathrm{gen}}$ is the MSE between predicted and target future Cosmos latents over valid views.

\subsection{MoSAIC: Mixture of Slotted Adapters}

As shown in Figure~\ref{fig:framework}, each of the $K$ clients trains a standard LoRA adapter for each layer on its private task-family dataset $D_k$. The server maintains $E=K>1$ complete expert branches and one selection router per adapted projection. A \emph{slot} identifies a client-associated LoRA adapter within a module, not a separate model.

\paragraph{Adapter Slotting.}
In round $\tau$, client $k$ uploads its adapter factors, which directly overwrite slot $k$ at each adapted module $m$:
\begin{equation}
    (A_{m,k}^{\tau,0},B_{m,k}^{\tau,0})
    \leftarrow
    (A_{m,k}^{\mathrm{loc},\tau},B_{m,k}^{\mathrm{loc},\tau}).
\end{equation}

No averaging occurs during installation. Client-to-slot identities remain fixed without restricting token assignments, and expert factors are refreshed each round, while selection routers persist only on the server. The protocol exchanges parameters, not raw trajectories.

\paragraph{Routed Expert Integration.}
For token representation $z$, the adapted projection is, omitting bias,
\begin{equation}
    h_m(z)=W_m^0z+\gamma_m\sum_{e=1}^{E}
    \pi_{m,e}(z)B_{m,e}A_{m,e}z,
    \qquad \gamma_m=\alpha_m/r,
\end{equation}
where $W_m^0$ is frozen and $\pi_m$ contains renormalized top-$k_{\mathrm{route}}$ weights, distinct from the dense probabilities pooled into $q_m^P$. The server jointly refines selection routers and both expert factors on mixed-task server data, starting from task-trained adapters. Path indices on weights are suppressed while each path retains its own parameters. Uploaded shared action-head weights are uniformly averaged across clients, refined at the server, and copied back to all clients.

\paragraph{Global Adapter Conversion.}
Let $\Delta W_{m,e}^{\mathrm{srv},\tau}=\gamma_m B_{m,e}^{\mathrm{srv},\tau}A_{m,e}^{\mathrm{srv},\tau}$ denote the complete server-refined adapter residual relative to the frozen backbone. PCEA converts these updates into a rank-constrained global adapter using detached routing evidence, as detailed below. This static conversion prepares client parameters rather than guaranteeing equivalence to the input-dependent MoE.

\paragraph{Personalized Adapter Redistribution.}
Client $k$ receives standard LoRA factors representing
\begin{equation}
    \Delta W_{m,k}^{\mathrm{init},\tau+1}
    =\Pi_r\!\left(
        \tfrac12\Delta W_{m,k}^{\mathrm{srv},\tau}
        +\tfrac12\Delta W_m^{\mathrm{glob},\tau}
    \right),
    \label{eq:personalized_adapter_redistribution}
\end{equation}
where $\Pi_r$ is a truncated SVD approximation of rank at most $r$. The client-specific component is the \emph{post-server expert}, not the original upload, which enables personalized parameter aggregation. 

\subsection{FARD: Foresight-to-Action Routing Distillation}

FARD uses understanding-generation agreement to guide action routing. For each valid module-sample pair $(m,b)$, define the geometric-mean teacher and its reliability:
\begin{equation}
\begin{aligned}
a_{m,b}&=\sum_e\sqrt{q^U_{m,b,e}q^G_{m,b,e}},
\qquad q^{UG}_{m,b}=\frac{\sqrt{q^U_{m,b}q^G_{m,b}}}{a_{m,b}},\\
r_{m,b}&=a_{m,b}\left[1-\frac{H(q^{UG}_{m,b})}{\log E}\right]_+.
\end{aligned}
\end{equation}

Thus, reliability combines path agreement with teacher concentration, so that unreliable routing signals are down-weighted during training. For valid samples $\mathcal{V}_m$, the loss is
\begin{equation}
\mathcal{L}_{\mathrm{FARD}}=\operatorname{mean}_{m}
\left[\frac{1}{|\mathcal{V}_m|}\sum_{b\in\mathcal{V}_m}
\operatorname{sg}(r_{m,b})
D_{\mathrm{JS}}\!\left(\operatorname{sg}(q^{UG}_{m,b}),q^A_{m,b}\right)\right],
\end{equation}
averaged over modules with nonempty $\mathcal{V}_m$ (zero if none). Here $\operatorname{sg}$ stops gradients while entropy and JS use natural logarithms, with numerical probability floors omitted for clarity. Gradients pass through the action-distribution argument, not the teacher or reliability, and upstream $U/G$ parameters can still receive indirect gradients through joint attention. The overall server objective therefore minimizes
\begin{equation}
\mathcal{L}_{\mathrm{MoSAIC}} = \mathcal{L}_{\mathrm{action}}
+ \lambda_{\mathrm{gen}}\mathcal{L}_{\mathrm{gen}}
+ \lambda_{\mathrm{aux}}\mathcal{L}_{\mathrm{aux}}
+ \lambda_{\mathrm{FARD}}^{(\tau)}\mathcal{L}_{\mathrm{FARD}},
\end{equation}
where $\mathcal{L}_{\mathrm{aux}}$ denotes the module-level auxiliary load-balancing loss averaged over the $E$ experts of each module, and $\lambda_{\mathrm{FARD}}^{(\tau)}$ follows the round-wise warmup schedule.

\subsection{PCEA: Path-Consensus Expert Aggregation}

PCEA weights complete expert updates rather than averaging their factors, since, in general,
\begin{equation}
\sum_e \omega_e B_eA_e \neq
\left(\sum_e \omega_e B_e\right)
\left(\sum_e \omega_e A_e\right).
\end{equation}

During server training, it collects detached, unnormalized Hellinger-barycenter evidence:
\begin{equation}
g_{m,b,e} = \left(
\frac{\sqrt{q^U_{m,b,e}}+\sqrt{q^G_{m,b,e}}+\sqrt{q^A_{m,b,e}}}{3}
\right)^2.
\end{equation}

Summing the valid sample occurrences across server steps and ranks within a round gives:
\begin{equation}
s_{m,e}=\sum_b g_{m,b,e},\qquad
w_{m,e}=\frac{s_{m,e}}{\sum_j s_{m,j}},\qquad
\bar a_m=\frac{\sum_e s_{m,e}}{N_m},
\end{equation}
where $N_m$ counts sample occurrences. The evidence retains its agreement mass rather than being normalized per sample. For observed module triplets $\mathcal{M}$, PCEA computes agreement-weighted global weights and a Hellinger midpoint:
\begin{equation}
    \bar w_e=\frac{\sum_{m\in\mathcal{M}}\bar a_m w_{m,e}}{\sum_{m\in\mathcal{M}}\bar a_m},
    \qquad
    \widetilde w_{m,e}=
    \frac{(\sqrt{w_{m,e}}+\sqrt{\bar w_e})^2}
    {\sum_j(\sqrt{w_{m,j}}+\sqrt{\bar w_j})^2}.
\end{equation}

Matched $U/G/A$ modules share these weights. Eligible unobserved modules use $\bar w$, and PCEA requires nonempty evidence. The unmatched state-projection adapter is aggregated uniformly.

For each physical module's own post-server factors, conversion gives:
\begin{equation}
    \Delta W_m^{\mathrm{glob},\tau}
    =\Pi_r\!\left(\sum_e\widetilde w_{m,e}\Delta W_{m,e}^{\mathrm{srv},\tau}\right).
    \label{eq:pcea_global_conversion}
\end{equation}
Implementation uses QR reduction on the stacked weighted factors, followed by a small SVD. FARD guides routing during server training, while PCEA consolidates the updates for client training.

\section{Experiments}
\label{sec:experiment}
\subsection{Experimental Settings}

\paragraph{Evaluation Protocol.} We evaluate \robofl \ under the task-silo protocol on RoboTwin 2.0~\citep{mu2025robotwin}, RLBench~\citep{james2020rlbench} and Franka robots\footnote{Full task settings and evaluation results are shown in the Appendix.}. We partition the 50 tasks into 8 client task-family groups and 1 server residual group for RoboTwin 2.0 \texttt{random}, and split the 8 RLBench tasks into 4 client partitions and 1 server partition, with 6 tasks distributed across the four clients and 2 tasks retained on the server. As for real-world applications, we partition 6 manipulation tasks into 4 clients and 1 server for Franka, creating an extreme \texttt{non-IID data scenario}.

\paragraph{Federated Optimization.}
All experiments initialize from the pretrained \textit{InternVLA-A1-3B} checkpoint, use full client participation, $100$ communication rounds, and one local epoch. Client and server steps are $500/200/300$ (RoboTwin / RLBench / Franka); client batch size is $32$, and server batch size is $8/16/16$. Both optimizers are AdamW with learning rate $1\times10^{-4}$, betas $(0.9,0.95)$, epsilon $10^{-8}$, weight decay $0.01$, and gradient-norm clipping at $1.0$, using cosine decay from $1\times10^{-4}$ to $1\times10^{-5}$ over $50{,}000/20{,}000/30{,}000$ decay steps and $2{,}500/400/600$ warmup steps.

\paragraph{MoSAIC Configuration.} MoSAIC uses eight expert branches per adapted module with top-$4$ server-side routing for RoboTwin 2.0, and four expert branches per adapted module with top-$2$ server-side routing for RLBench and the real-world applications. Each expert uses rank $r=16$, LoRA parameter $\alpha=32$ (effective scaling $\alpha/r=2$), and dropout $0.05$. The generation and load-balancing coefficients are $\lambda_{\mathrm{gen}}=0.01$ and $\lambda_{\mathrm{aux}}=0.001$. FARD uses $\lambda_{\mathrm{FARD}}=0.01$ with a $10$-round warmup. 

\paragraph{Baselines.} We compare against \textbf{InternVLA-A1}~\citep{cai2026internvla} and \textbf{Motus}~\citep{bi2026motus}, which are centralized PEFT references; \textbf{FedAvg}~\citep{mcmahan2017communication} performs standard weight averaging; \textbf{FedMoE}~\citep{mei2024fedmoe} uses client-specific expert sub-MoEs; \textbf{FedVLA}~\citep{miao2025fedvla} combines instruction-oriented scene parsing, dual-gating MoE, and expert-driven aggregation; and \textbf{ForgeVLA}~\citep{zhou2026forgevla} targets heterogeneous federated robot data with weak or missing language supervision. We adapt them to the same FL-PEFT with Kaiming-random-initialized adapter.

\begin{table*}[t]
    \centering
    \caption{RoboTwin 2.0 success rates on a 25-task challenging subset under random condition (selected as tasks where FedAvg $<$ 80\%). Overall reports performance on all 50 tasks.}
    \label{tab:robottwin_remaining_tasks}
    \setlength{\tabcolsep}{1pt}
    \renewcommand{\arraystretch}{1.22}
    \resizebox{1\textwidth}{!}{%
    \begin{tabular}{cccccccccccccc}
    \toprule
    \rowcolor{roboflblue}
    \color{white}{\textbf{Method}} &
    \color{white}{\taskname{beat}{ham.}} &
    \color{white}{\taskname{rank}{size}} &
    \color{white}{\taskname{hand}{blo.}} &
    \color{white}{\taskname{hang}{mug}} &
    \color{white}{\taskname{lift}{pot}} &
    \color{white}{\taskname{move}{can}} &
    \color{white}{\taskname{move}{sta.}} &
    \color{white}{\taskname{pick}{div.}} &
    \color{white}{\taskname{pick}{dual}} &
    \color{white}{\taskname{place}{L}} &
    \color{white}{\taskname{place}{R}} &
    \color{white}{\taskname{bread}{ski.}} &
    \color{white}{\taskname{can}{basket}}\\
    \midrule
        \multicolumn{14}{c}{\texttt{CENTRALIZED TRAINING}} \\
    \midrule
    \rowcolor{roboflgray}
    \textsc{InternVLA} & 73 & 78 & \textit{62} & 27 & 32 & 76 & \textbf{53} & 68 & 74 & \textit{85} & \textbf{84} & \textit{79} & 60 \\
    \textsc{Motus} & \textbf{82} & \textit{86} & 60 & 20 & \textbf{40} & \textit{86} & 48 & \textit{70} & 80 & \textit{85} & 80 & \textbf{80} & 69 \\
    \midrule
    \multicolumn{14}{c}{\texttt{FEDERATED LEARNING}\tiny\textsc{(LoRA/MoE)}} \\
    \midrule
    \rowcolor{roboflgray}
    \textsc{FedAvg} & 75 & 74 & 52 & 24 & 33 & 67 & 39 & 69 & 77 & 79 & 79 & 74 & 72 \\
    \textsc{FedMoE} & 69 & 71 & 42 & 21 & \textit{34} & 57 & 34 & 61 & 70 & 84 & 80 & 66 & 64 \\
    \midrule
    \rowcolor{roboflgray}
    \textsc{ForgeVLA} & 75 & 83 & 57 & \textbf{35} & \textbf{40} & 83 & 44 & 67 & \textit{82} & \textbf{87} & 82 & 78 & 68 \\
    \textsc{FedVLA} & 73 & 85 & 61 & 26 & 33 & 85 & 45 & 62 & \textbf{84} & 80 & 82 & 77 & \textit{74} \\
    \rowcolor{roboflgreen}
    \textbf{\robofl} & \textit{78} & \textbf{92} & \textbf{73} & \textit{30} & 24 & \textbf{87} & \textit{50} & \textbf{72} & \textbf{84} & \textbf{87} & \textit{83} & \textbf{80} & \textbf{82} \\
    \midrule
    \rowcolor{roboflblue}
    \color{white}{\textbf{Method}} &
    \color{white}{\taskname{dual}{shoe}} &
    \color{white}{\taskname{mouse}{pad}} &
    \color{white}{\taskname{place}{bas.}} &
    \color{white}{\taskname{place}{sca.}} &
    \color{white}{\taskname{press}{sta.}} &
    \color{white}{\taskname{bottles}{dus.}} &
    \color{white}{\taskname{put}{cab.}} &
    \color{white}{\taskname{rotate}{QR}} &
    \color{white}{\taskname{scan}{obj}} &
    \color{white}{\taskname{stack}{bow.}} &
    \color{white}{\taskname{stamp}{seal}} &
    \color{white}{\taskname{turn}{swi.}} &
    \color{white}{\textbf{Overall}} \\
    \midrule
        \multicolumn{14}{c}{\texttt{CENTRALIZED TRAINING}} \\
    \midrule
    \rowcolor{roboflgray}
    \textsc{InternVLA} & 64 & \textbf{70} & \textbf{82} & \textbf{81} & 74 & \textbf{90} & \textbf{68} & 77 & 64 & \textbf{86} & \textit{68} & 43 & \textit{81.60} \\
    \textsc{Motus} & \textit{73} & \textit{64} & 79 & 73 & 72 & \textit{89} & \textit{67} & \textbf{87} & 65 & 77 & 66 & 45 & 80.96 \\
    \midrule
        \multicolumn{14}{c}{\texttt{FEDERATED LEARNING}\tiny\textsc{(LoRA/MoE)}} \\
    \midrule
    \rowcolor{roboflgray}
    \textsc{FedAvg} & 67 & 56 & 79 & 75 & \textit{75} & 77 & \textbf{68} & 79 & 58 & 79 & 57 & 48 & 78.92 \\
    \textsc{FedMoE} & 60 & 54 & 75 & 69 & 74 & 79 & 62 & 73 & 31 & 78 & 60 & 38 & 75.12 \\
     \midrule
     \rowcolor{roboflgray}
    \textsc{ForgeVLA} & 70 & 59 & \textit{80} & \textit{77} & \textbf{78} & 83 & 61 & 77 & 69 & 75 & 59 & \textit{52} & 80.70 \\
    \textsc{FedVLA} & 71 & 53 & 70 & 75 & \textit{75} & 84 & 60 & 75 & \textit{70} & \textit{80} & 65 & 51 & 79.32 \\
    \rowcolor{roboflgreen}
    \textbf{\robofl} & \textbf{74} & 58 & 72 & 74 & 74 & \textbf{90} & \textbf{68} & \textit{81} & \textbf{74} & \textbf{86} & \textbf{70} & \textbf{53} & \textbf{83.12} \\
    \bottomrule
    \end{tabular}
    }
\vspace{-1.5em}
\end{table*}

\subsection{Quantitative Results for Simulation}

\paragraph{Model Performance on RoboTwin 2.0.} Table~\ref{tab:robottwin_remaining_tasks} reports the 25 RoboTwin 2.0 tasks on which FedAvg stays below 80\%, since the remaining tasks are near-saturated and thus uninformative. \robofl \ attains the highest overall success rate of 83.12\%, exceeding FedAvg by 4.20\% and also surpassing centralized PEFT with InternVLA-A1 and Motus. This pattern matches \mosaic's design: retaining locally specialized adapters as expert branches and learning their input-dependent composition rather than immediately averaging them. Best or joint-best results on 15 displayed tasks show that the gains extend across tasks. Overall, \robofl \ delivers the strongest performance in terms of single-adapter client cost, demonstrating that server-side expert assembly can outperform both federated averaging and centralized PEFT across heterogeneous task silos.

\begin{wraptable}{r}{0.55\textwidth}
\centering
\vspace{-2.5em}
\caption{Average success rates (\%) for eight RLBench tasks with 4-expert MoSAIC for \robofl.}
\label{tab:rlbench}
\setlength{\tabcolsep}{3pt}
\renewcommand{\arraystretch}{1.15}
\resizebox{0.55\textwidth}{!}{%
\begin{tabular}{ccc|ccc}
\toprule
\rowcolor{roboflblue}
\color{white}{\textbf{Method}} & \color{white}{\textbf{InternVLA}} & \color{white}{\textbf{Motus}} & \color{white}{\textbf{ForgeVLA}} & \color{white}{\textbf{FedVLA}} & \color{white}{\textbf{\robofl}} \\
\midrule
Type &   \multicolumn{2}{c|}{\texttt{Central-LoRA}} & \multicolumn{1}{c}{\texttt{FL-LoRA}} &  \multicolumn{1}{c}{\texttt{FL-MoE}} & \multicolumn{1}{c}{\texttt{LoRA$\rightarrow$MoE}}\\
\midrule
\rowcolor{roboflgray}
\textsc{Success} & 52.75 & 52.50 & 34.25 & 31.00 & 41.25 \\
\bottomrule
\end{tabular}
}
\vspace{-1.5em}
\end{wraptable}
\paragraph{Model Performance on RLBench.} We further examine how \robofl performs on RLBench with only eight tasks under a 4-client, 4-expert MoSAIC in Table~\ref{tab:rlbench}. \robofl \ remains the best FL method at 41.25\%, yet the margin over centralized PEFT reverses relative to RoboTwin, where \robofl \ also beats InternVLA-A1 and Motus. With fewer tasks and clients, each expert sees less complementary evidence for routed composition, so the advantage of structured expert assembly is smaller. These results suggest that \robofl \ is most effective in large-scale, task-heterogeneous regimes rather than in small, low-data suites.

\begin{wraptable}{r}{0.55\textwidth}
\centering
\vspace{-1em}
\caption{Client resources and logical communication on RoboTwin 2.0. Communication combines upload and download per participating client per round.}
\label{tab:client_resources}
\setlength{\tabcolsep}{4pt}
\renewcommand{\arraystretch}{1.15}
\resizebox{0.55\textwidth}{!}{%
\begin{tabular}{cccccc}
\toprule
\rowcolor{roboflblue}
{\color{white}\textbf{Method}} &
{\color{white}\shortstack[c]{Adapters\\c / s}} &
{\color{white}\shortstack[c]{Params\\(M)}} &
{\color{white}\shortstack[c]{Comm.\\MiB}} &
{\color{white}\shortstack[c]{Mem.\\GiB}} &
{\color{white}\shortstack[c]{Success\\(\%)}} \\
\midrule
\rowcolor{roboflgray}
\textsc{FedAvg} & 1 / 1 & 40.85 & 311.63 & 8.33 & 78.92 \\
\rowcolor{roboflgray}
\textsc{FedMoE} & 4 / 8 & 158.11 & 1,206.28 & 11.44 & 75.12 \\
\textsc{ForgeVLA} & 1 / 1 & 40.85 & 312.06 & 8.33 & 80.70 \\
\rowcolor{roboflgray}
\textsc{FedVLA} & 8 / 8 & 312.93 & 2,362.85 & 15.59 & 79.32 \\
\rowcolor{roboflgreen}
\textbf{\robofl} & \textbf{1 / 8} & \textbf{40.85} & \textbf{311.63} & \textbf{8.33} & \textbf{83.12} \\
\bottomrule
\end{tabular}
}
\vspace{-1em}
\end{wraptable}
\paragraph{Client Resources and Communication.}
Table~\ref{tab:client_resources} compares adaptation resources at the client boundary, combining upload and download into a per-round communication total. With one rank-16 adapter per client, \robofl \ matches FedAvg's $40.85$M trainable parameters and $311.63$~MiB payload, while retaining eight server-side expert branches. Relative to MoE-based FL methods FedMoE and FedVLA, client parameters decrease by up to 86.95\% and per-round communication by up to 86.81\%, while RoboTwin 2.0's success rates increase by 8 and 3.8 percentage points, respectively. Peak allocated GPU memory falls from $15.59$ to $8.33$~GiB in the standardized client-architecture microbenchmark. These results demonstrate that \robofl \ achieves the best accuracy-efficiency trade-off: it attains the highest success rate with the same computational and communication overhead as single-adapter baselines, while remaining far below client-side MoE.

\subsection{Ablation Studies}

\par 
\begin{wrapfigure}{r}{0.55\textwidth}
    \centering
    \vspace{-1.2em}
    \includegraphics[width=\linewidth]{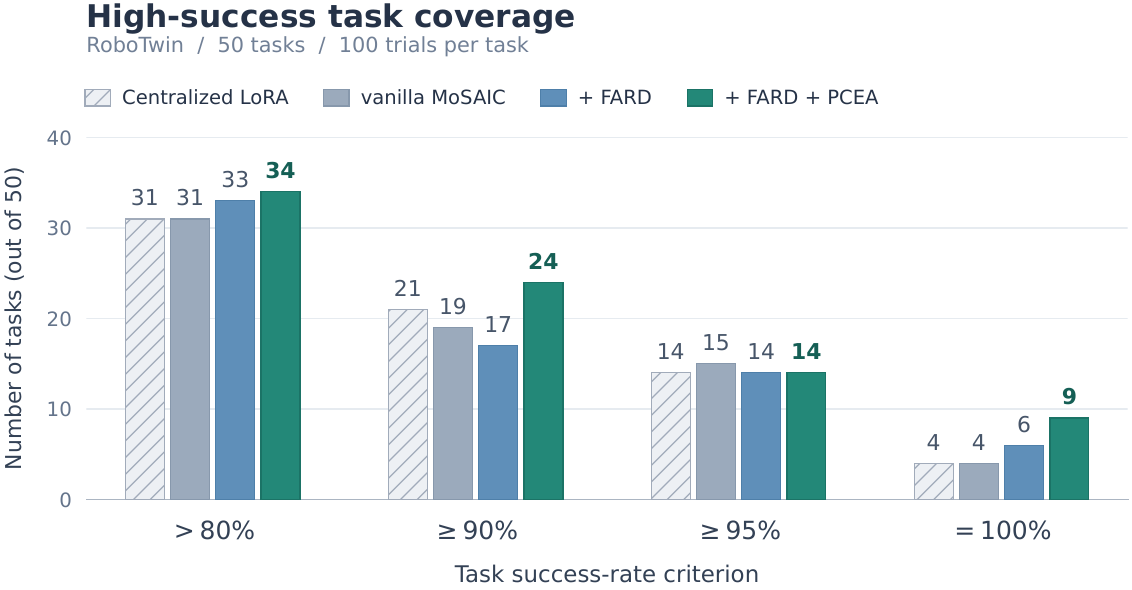}
    \vspace{-2em}
    \caption{Success-threshold coverage on RoboTwin. Bars count tasks satisfying each success-rate criterion across 50 tasks with 100 trials per task.}
    \vspace{-1em}
    \label{fig:success_threshold_coverage}

\end{wrapfigure}

\paragraph{Influence of FARD and PCEA.}
\label{sec:ablation}
We present a sequential ablation of FARD and PCEA, using vanilla MoSAIC as the baseline and centralized LoRA as an external reference under its own training schedule with RoboTwin 2.0, as shown in Figure~\ref{fig:success_threshold_coverage}. 
We evaluate PCEA on top of FARD because its consensus-weighted conversion is designed to exploit the routing agreement.
FARD alone increases the number of tasks exceeding $80\%$ success from $31$ to $33$, but reduces coverage at $\geq90\%$ from $19$ to $17$. Adding PCEA raises these counts to $34$ and $24$, surpassing both vanilla MoSAIC and centralized LoRA at the latter threshold. Tasks with 100/100 successes likewise increase from $4$ for both references to $6$ with FARD and $9$ with FARD+PCEA. At $\geq95\%$, however, both variants cover $14$ tasks versus $15$ for vanilla MoSAIC. These results support combining routing distillation with expert conversion, potentially preserving coordinated task knowledge across federated rounds and supporting more consistent execution and broader coverage of high-success tasks.

\paragraph{Routing Coupling and Intervention.}
\label{sec:routing_probe}

\begin{wrapfigure}{r}{0.55\textwidth}
    \centering
    \vspace{-1em}
    \includegraphics[width=\linewidth]{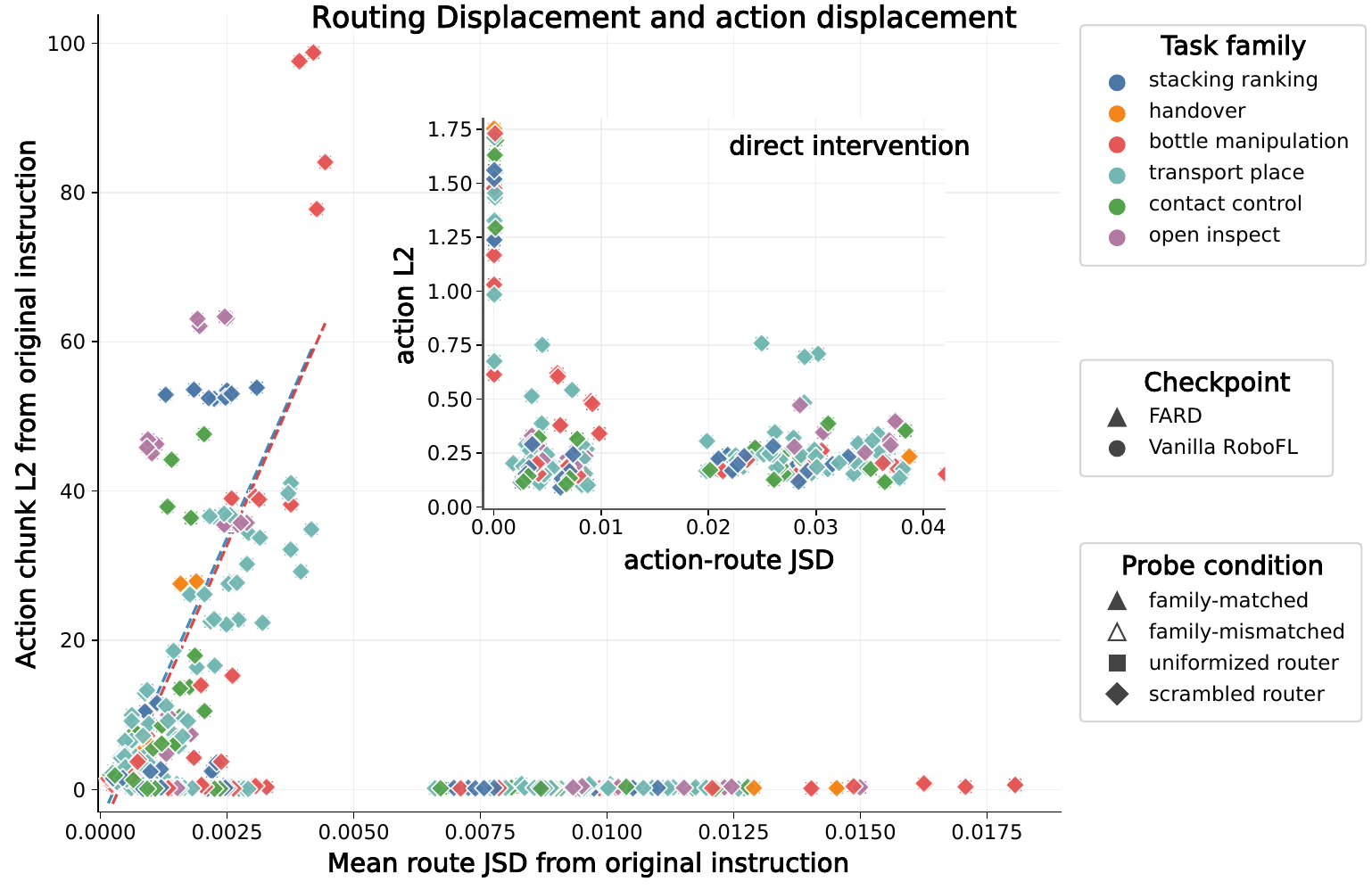}
    \vspace{-2em}
    \caption{Route-to-action coupling under instruction swaps and direct action-router interventions. The inset magnifies the direct intervention regime.}
    \label{fig:route_action_coupling}
    \vspace{-1em}
\end{wrapfigure}

We probe route-to-action coupling across 50 RoboTwin 2.0 tasks while holding the initial observation and robot state fixed. Instruction substitution changes only the language condition, whereas uniformized and scrambled controls intervene only on the action-router distribution. As shown in Figure~\ref{fig:route_action_coupling}, instruction swaps produce strong pooled route-to-action coupling, establishing that task semantics reach action generation through routing.
\par 
The direct intervention isolates the action router more strictly. Scrambling produces a larger routing displacement for FARD, yet a smaller action displacement than vanilla \robofl. Thus, FARD makes routing more selective without making the resulting policy brittle. Together, the probes support FARD as a cross-path routing-alignment mechanism, while PCEA performs the corresponding path-consistent expert conversion.

\subsection{Real-world Applications with Franka}

\begin{table*}[t]
    \centering
    \caption{Real-world applications with six single-arm Franka tasks SR (\%) with three replicates.}
    \label{tab:franka_real_world}
    \renewcommand{\taskname}[2]{{\ttfamily\makecell[c]{#1\\#2}}}
    \setlength{\tabcolsep}{6pt}
    \renewcommand{\arraystretch}{1.28}
    \resizebox{1\textwidth}{!}{%
    \begin{tabular}{cccccccc|c}
\toprule
\rowcolor{roboflblue}
    \color{white}\textbf{Method} &
    \color{white}\textbf{Type} &
    \color{white}\taskname{adjust}{bottle} &
    \color{white}\taskname{stamp}{seal} &
    \color{white}\taskname{stack}{cups} &
    \color{white}\taskname{screw}{bottle} &
    \color{white}\taskname{pour}{water} &
    \color{white}\taskname{test}{tube} &
    \cellcolor{roboflblue}%
    {\color{white}\shortstack[c]{%
        \textbf{Mean}
        $\pm$ \textbf{Avg. SD}%
    }} \\
\midrule

    \rowcolor{roboflgray}
    \textsc{InternVLA} & \textsc{Centralized-LoRA} &
    63.33 & 53.33 & 41.67 &
    31.67 & 48.33 & 43.33 &
    $46.94 \tiny \pm 7.80$ \\
    \midrule

    \textsc{ForgeVLA} & \textsc{FL-LoRA} &
    41.67 & 31.67 & 46.67 &
    28.33 & 31.67 & 33.33 &
    $35.56 \tiny \pm 7.03$ \\

    \textsc{FedVLA} & \textsc{FL-MoE} &
    43.33 & 41.67 & 33.33 &
    13.33 & 40.00 & 30.00 &
    $33.61 \tiny \pm 5.60$ \\

    \rowcolor{roboflgreen}
    \textbf{\robofl} & \textsc{FL-LoRA}$\rightarrow$\textsc{MoE} &
    \textbf{68.33} & \textbf{66.67} & \textbf{68.33} &
    \textbf{36.67} & \textbf{56.67} & \textbf{58.33} &
    $\mathbf{59.17 \tiny \pm 7.61}$ \\
    \bottomrule
    \end{tabular}%
    }
\end{table*}

\begin{figure*}[t]
    \centering
    \vspace{-1em}
    \includegraphics[width=0.95\textwidth]{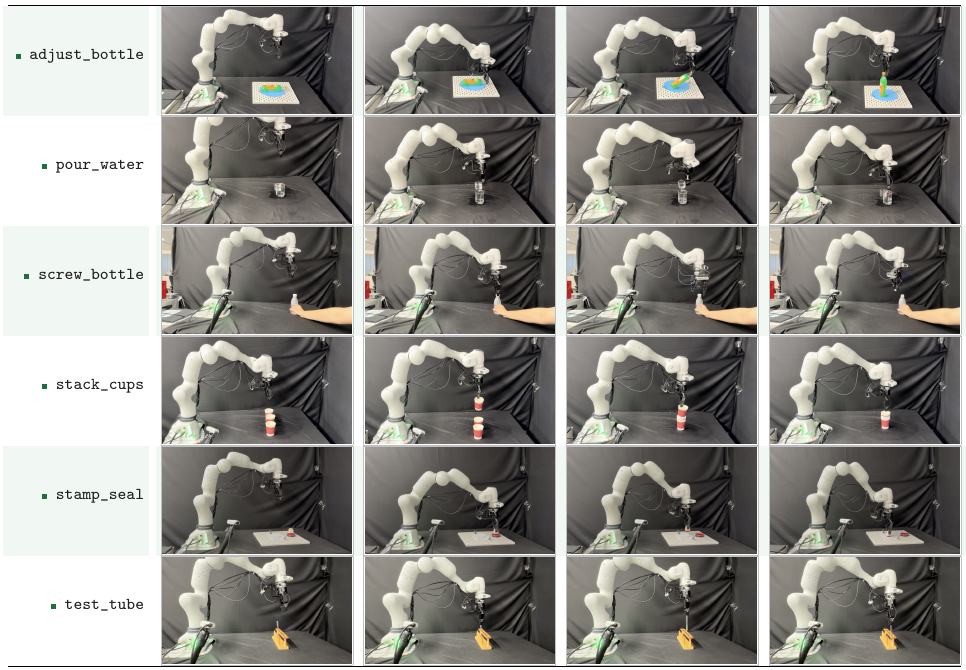}
    \vspace{-1em}
    \caption{Qualitative successful real-world Franka rollouts for each of six single-arm Franka tasks.}
    \label{fig:franka_success}
    \vspace{-1em}
\end{figure*}

\vspace{-0.5em}
\paragraph{Quantitative Analysis.} Table~\ref{tab:franka_real_world} reports success rates on six single-arm Franka tasks. We report mean success rates over three evaluation rounds of 20 trials per task. \robofl \ attains the highest Overall success rate of 59.17\%, exceeding centralized InternVLA by 12.23\% and federated ForgeVLA and FedVLA by 23.61\% and 25.56\%. MoSAIC therefore extends beyond simulation: installing each client's single LoRA as a routed server expert, rather than averaging updates, preserves input-dependent specialization under real perception, contact, and hardware variability, even surpassing pooled centralized PEFT at the cost of a single adapter per client.

\vspace{-0.5em}
\paragraph{Qualitative Analysis.}
Figure~\ref{fig:franka_success} shows successful rollouts on six Franka tasks. Execution follows the simulation's three-phase structure: approach, sustained contact, and sequenced state change. This closed-loop coherence reflects the method's design: task-trained experts remain distinct under server routing, while understanding-generation consensus guides action routing, so contact-rich behaviors remain input-dependent rather than averaging into generic motions. For example, \textit{adjust bottle} task seats rather than strikes, \textit{screw bottle} task sustains rotation through the cap.

\vspace{-1em}
\section{Conclusion}
\vspace{-0.5em}
We presented \robofl, a federated framework for learning world-action models from task-siloed robot data without centralizing raw trajectories. \mosaic \ installs locally trained LoRA adapters as server MoE experts and jointly refines routing and expert parameters. Specifically, \fard \ aligns action routing with understanding-generation consensus and \pcea \ consolidates expert updates into a rank-constrained global adapter for personalized redistribution. Across simulation and real-robot experiments, their combination delivers the highest overall success rates among the evaluated federated methods while keeping client training at single-adapter cost.


\section*{AI Use Statement}
The authors used AI-based writing and coding assistants to help draft and polish portions of the text and to support the implementation and analysis scripts. All problem formulation, method design, experimental execution, data collection, and interpretation of results were carried out and verified by the authors, who take full responsibility for the content of this paper.

\section*{Ethics Statement}
This work studies federated learning of world-action models for robotic manipulation. Experiments in simulation use the publicly released RoboTwin 2.0 and RLBench benchmarks, and the real-robot evaluations collect no personally identifiable information. The motivating concern of the paper is that robot trajectories can reveal private environments, user routines, or proprietary procedures, and that centralized pooling of such data may be undesirable. The proposed protocol keeps raw trajectories at each institution and exchanges only adaptation parameters, thereby reducing this exposure; however, we do not claim formal privacy guarantees, such as differential privacy or secure aggregation, and parameter updates may still leak information in adversarial settings. We therefore position the method as an architectural alternative to raw-data centralization rather than as a complete privacy solution. We also note that broad manipulation policies can be misused, and we follow the ICLR Code of Ethics in conducting and reporting this work.

\section*{Reproducibility Statement}
We summarize the resources provided to support reproducibility. The full set of training and evaluation settings appears in Table~\ref{tab:benchmark_hyperparameters} and Appendix~\ref{app:experimental_details}; the task-silo assignments for the eight RoboTwin 2.0 clients and the four RLBench clients are listed in Tables~\ref{tab:task_silo_manifest} and~\ref{tab:rlbench_task_silo_manifest}; and the observation, action, and optimization interfaces are specified in the same appendix. The client resources, logical communication accounting, and the architecture profiles behind the memory measurements are documented in Appendix~\ref{app:common_parameter_exchange}. The MoSAIC, FARD, and PCEA formulations, including the adapter slotting, router integration, personalized redistribution, and consensus-weighting rules, are given in the Method section and implemented as described in Appendix~\ref{app:common_parameter_exchange}. We plan to release the source code, launch configurations, and evaluation scripts as supplementary materials; these will include data preprocessing and normalization steps, evaluation packages for both benchmarks, and scripts used to generate the resource and ablation figures. Simulation results use a single fixed evaluation seed; real-robot results use a fixed scripted protocol with 20 trials per task. Differences below a few points, and per-task swings on the 8-task RLBench suite, should be read as descriptive rather than statistically established.

\section*{Acknowledgment}
This work was supported by the National Natural Science Foundation of China (62476011) and (625B2090), the Beijing Natural Science Foundation (L252060).

\bibliography{iclr2027_conference}
\bibliographystyle{iclr2027_conference}

\clearpage

\appendix
\section*{Appendix}
\section{Additional Experimental Details for Simulations}
\label{app:experimental_details}

\subsection{Shared Implementation Details}
\label{app:common_parameter_exchange}

\paragraph{Parameter exchange.} Both benchmarks use the same adapter-based implementation. LoRA adapts the attention, MLP, and state projections. Clients upload complete LoRA factors together with the trainable action input/output projections and action-time MLP weights, while frozen backbone weights are excluded. Shared action-head weights are averaged uniformly and synchronized separately from expert conversion. Server MoE checkpoints are saved before global adapter conversion and personalized redistribution.

\label{app:task_silo_manifest}

\paragraph{Server participation and shared-compute assumption.}
\robofl\ lets the central server take part in training rather than act only as an aggregator. We regard this as a reasonable and common use of the decentralized setting: the server is typically the best-provisioned node, and federated methods routinely exploit server-side computation, for example through server-level adaptive optimization~\citep{reddi2021adaptive}, distillation of the client ensemble on server or surrogate data~\citep{lin2020feddf,chen2021fedbe}, or data-free server-side generators~\citep{zhu2021datafree}. Under this assumption, the server holds one residual task group and performs the routed expert-refinement stage on it, which enables a single shared router over task-trained experts; the alternative reading, in which the server has no data of its own, would leave the router without a common distribution to learn from. Crucially, this does not amount to centralized training: the server group covers only a small fraction of each benchmark ($7/50$, $2/8$, and $2/6$ tasks on RoboTwin~2.0, RLBench, and Franka), so the bulk of the task distribution still resides with the clients. The server is also not given a disproportionate optimization budget: its per-round steps match the per-client local steps ($500/200/300$ for the three scenarios), and its task group is comparable in size to the client partitions ($3$-$8$, $1$-$2$, and $1$ task per client). Clients therefore continue to train a single adapter, and the additional routing and multi-expert computation remains on the server.

\paragraph{Comparison Fairness with Centralized Training Baselines}
We treat centralized PEFT as a deliberately \emph{stronger} reference than any federated method, not as an equal-budget competitor. A centralized run observes the union of all task families in a single optimizer, with no task silos, no non-IID client drift, and no communication rounds. That \robofl \ matches or exceeds this reference on RoboTwin~2.0 and Franka while operating under task-silo federated learning, and, consistent with its mechanism, trails it only on the low-data RLBench suite, where less complementary evidence exists for expert composition, is the intended empirical claim of Section~\ref{sec:experiment}.

The comparison is also capacity-matched at the point of deployment. The entity that is trained, redistributed, and evaluated is a \emph{single} rank-16 LoRA adapter per client, identical in family, rank, and scaling to the adapters used by federated averaging and by the centralized PEFT references. The eight-expert server MoE is a training-time construct that never leaves the server; clients never hold routers or multiple experts, and the deployed policy is the redistributed single adapter, not the server MoE. At the client boundary, \robofl \ is therefore indistinguishable from single-adapter FL in trainable parameters, communication, and memory (Table~\ref{tab:client_resources}), and is far below client-side MoE methods. The reported gains consequently cannot be attributed to a larger deployed model or to additional client-side capacity: they follow from \emph{how} task-specialized adapters are composed on the server under heterogeneity, not from giving the learned policy more parameters than its baselines. Notably, this advantage is obtained under an extreme non-IID, task-disjoint regime in which every client still trains and communicates a single adapter.

\paragraph{Configuration.} Table~\ref{tab:benchmark_hyperparameters} summarizes the numerical settings. RLBench training entries marked $\dagger$ are launcher defaults, not reconstructed settings of the reported checkpoints. Task assignments and evaluation semantics are specified separately below. Last but not least, all the FL-based baselines follow the same training procedure as \robofl.

\begin{table*}[t]
    \centering
    \caption{Training and evaluation settings for the simulation and real-world scenarios. Execution horizon is specified independently of prediction length.}
    \vspace{0.5em}
    \label{tab:benchmark_hyperparameters}
    \setlength{\tabcolsep}{11pt}
    \renewcommand{\arraystretch}{1.15}
    \resizebox{\textwidth}{!}{%
    \begin{tabular}{@{}>{\raggedright\arraybackslash}p{0.21\textwidth}>{\raggedright\arraybackslash}p{0.245\textwidth}>{\raggedright\arraybackslash}p{0.245\textwidth}>{\raggedright\arraybackslash}p{0.245\textwidth}@{}}
    \toprule
    Setting & RoboTwin 2.0 & RLBench & Franka \\
    \midrule
    \multicolumn{4}{c}{\texttt{Optimization and adapters}} \\
    \midrule
    Pretrained checkpoint & \multicolumn{3}{c}{InternVLA-A1-3B} \\
    Clients / experts & $8\,/\,8$ & $4\,/\,4$ & $4\,/\,4$ \\
    Routing top-$k$ & $4$ & $2$ & $2$ \\
    Communi. rounds & $100$ & $100$ & $100$ \\
    Local steps / round & $500$ & $200$ & $300$ \\
    Local epochs & $1$ & $1$ & $1$ \\
    Server steps / round & $500$ & $200$ & $300$ \\
    Peak / final LR & $10^{-4}\,/\,10^{-5}$ & $10^{-4}\,/\,10^{-5}$ & $10^{-4}\,/\,10^{-5}$ \\
    Warmup / decay steps & $2{,}500\,/\,50{,}000$ & $400\,/\,20{,}000$ & $600\,/\,30{,}000$ \\
    LoRA rank / $\alpha$ & $16\,/\,32$ & $16\,/\,32$ & $16\,/\,32$ \\
    LoRA dropout & $0.05$ & $0.05$ & $0.05$ \\
    $\lambda_{\mathrm{gen}}\,/\,\lambda_{\mathrm{aux}}$ & $0.01\,/\,0.001$ & $0.01\,/\,0.001$ & $0.01\,/\,0.001$ \\
    \midrule
    \multicolumn{4}{c}{\texttt{Observation and action interface}} \\
    \midrule
    Views & Head + two wrists & Front + two masked slots & External stand + wrist + masked slot \\
    Resolution & $224\times224$ & $224\times224$ & $224\times224$ \\
    State & Joint configuration & XYZ + Euler + gripper & Joint + gripper \\
    Action & Joint-space offsets & $\Delta$XYZ + absolute Euler + gripper & Absolute joint targets \\
    History stride & $15$ & $1$ & $15$ \\
    Predicted chunk & $50$ & $8$ & $50$ \\
    Inference steps & $10$ & $10$ & $10$ \\
    \midrule
    \multicolumn{4}{c}{\texttt{Evaluation protocols}} \\
    \midrule
    Evaluated tasks & $50$ & $8$ & $6$ \\
    Trials / task & $100$ & $50$ & $20$ \\
    Executed actions & $30$ & $8$ & $50$ \\
    Initialization & Randomized & seed 42 & Operator reset \\
    \bottomrule
    \end{tabular}
    }
\end{table*}

\subsection{Hyperparameter and Design Choices}
\label{app:hyperparameters}

The non-obvious hyperparameters follow standard practice. The personalization coefficient $\tfrac12$ in Eq.~\ref{eq:personalized_adapter_redistribution} represents an equal global-local model interpolation, a standard personalization strategy with generalization guarantees~\citep{mansour2020three}. The local term preserves client specialization, while the global term shares server-acquired knowledge, and fixing it avoids introducing a per-client hyperparameter. The auxiliary weights are kept small because $\mathcal{L}_{\mathrm{action}}$ remains the primary objective and only the magnitude of such weights matters: we use $\lambda_{\mathrm{aux}}=0.001$~\citep{kendall2018multi}, with the load-balancing scale following the sparse-MoE convention~\citep{shazeer2017sparsely,fedus2022switch}. FARD is likewise a routing regularizer based on detached Jensen-Shannon distillation~\citep{hinton2015distilling}, so $\lambda_{\mathrm{FARD}}=0.01$, and it is warmed up over $10$ rounds because the reliability weight is uninformative before the action router stabilizes. Warmup is a standard variance-reduction device~\citep{liu2020variance}. 

\subsection{RoboTwin 2.0}
\label{app:robottwin_task_silo_manifest}

\paragraph{Task-silo manifest.} Table~\ref{tab:task_silo_manifest} lists the task-disjoint client and server assignments implemented by \texttt{fedforesight\_category}. Episodes remain within their assigned task partition, while the server residual group is used for routed optimization.

\paragraph{Observation and action interface.} As shown in the Table~\ref{tab:benchmark_hyperparameters}, predicted joint-space offsets are converted to joint targets using the configuration at the start of each action queue. The evaluator truncates predictions using its execution-horizon argument, while prediction length does not determine the number of executed actions.

\begin{table}[t]
    \centering
    \caption{Task assignments in the explicit eight-client task-silo implementation. Counts refer to tasks, not episodes or training frames.}
    \vspace{0.5em}
    \label{tab:task_silo_manifest}
    \setlength{\tabcolsep}{20pt}
    \renewcommand{\arraystretch}{1.15}
    \resizebox{1\textwidth}{!}{%
    \begin{tabular}{@{}lr>{\raggedright\arraybackslash}p{0.75\textwidth}@{}}
    \toprule
    Partition & Count & Tasks \\
    \midrule
    Client 0 & 5 & \texttt{grab\_roller}, \texttt{pick\_diverse\_bottles}, \texttt{pick\_dual\_bottles}, \texttt{put\_bottles\_dustbin}, \texttt{put\_object\_cabinet} \\
    Client 1 & 7 & \texttt{place\_object\_basket}, \texttt{place\_bread\_basket}, \texttt{place\_bread\_skillet}, \texttt{place\_can\_basket}, \texttt{place\_cans\_plasticbox}, \texttt{place\_container\_plate}, \texttt{place\_empty\_cup} \\
    Client 2 & 8 & \texttt{place\_a2b\_left}, \texttt{place\_a2b\_right}, \texttt{place\_fan}, \texttt{place\_mouse\_pad}, \texttt{place\_object\_scale}, \texttt{place\_object\_stand}, \texttt{place\_phone\_stand}, \texttt{place\_shoe} \\
    Client 3 & 6 & \texttt{stack\_blocks\_three}, \texttt{stack\_blocks\_two}, \texttt{stack\_bowls\_three}, \texttt{stack\_bowls\_two}, \texttt{blocks\_ranking\_rgb}, \texttt{blocks\_ranking\_size} \\
    Client 4 & 5 & \texttt{adjust\_bottle}, \texttt{lift\_pot}, \texttt{move\_pillbottle\_pad}, \texttt{move\_playingcard\_away}, \texttt{move\_stapler\_pad} \\
    Client 5 & 3 & \texttt{beat\_block\_hammer}, \texttt{press\_stapler}, \texttt{stamp\_seal} \\
    Client 6 & 6 & \texttt{click\_alarmclock}, \texttt{click\_bell}, \texttt{turn\_switch}, \texttt{open\_laptop}, \texttt{open\_microwave}, \texttt{rotate\_qrcode} \\
    Client 7 & 3 & \texttt{scan\_object}, \texttt{shake\_bottle}, \texttt{shake\_bottle\_horizontally} \\
    \midrule
    Server & 7 & \texttt{place\_burger\_fries}, \texttt{place\_dual\_shoes}, \texttt{dump\_bin\_bigbin}, \texttt{handover\_block}, \texttt{handover\_mic}, \texttt{hanging\_mug}, \texttt{move\_can\_pot} \\
    \bottomrule
    \end{tabular}
    }
\end{table}

\paragraph{Sampling and update accounting.} Local training cycles through a shuffled frame-level loader, without task-balanced sampling. The epoch setting repeats the local-step loop rather than guaranteeing a full dataset traversal of the dataset. For $C$ clients, $K$ rounds, $L$ local steps, $E$ repetitions, and $M$ server steps, the update counts are $KLE$ per client, $CKLE$ across clients, and $KM$ at the server. The schedule is given in Table~\ref{tab:benchmark_hyperparameters}.

\paragraph{Batch-size conventions.} Client batch size is per local update, while server batch size is per distributed rank. With $R$ server ranks and per-rank batch size $B_{\mathrm{srv}}$, the effective server batch is $R B_{\mathrm{srv}}$. Client optimizer states are retained separately across rounds, while server optimizer state persists across expert replacement.

\paragraph{Randomized initialization.} As shown in Table~\ref{tab:benchmark_hyperparameters}, unstable or expert-unsuccessful candidate scenes are excluded before policy evaluation. Accepted trials therefore need not correspond to consecutive candidate seeds. The evaluator defaults to the \texttt{unseen} instruction split.

\begin{table*}[t]
    \centering
    \caption{RoboTwin 2.0 task success rates (\%) on the complementary 25 tasks not shown in Table~\ref{tab:robottwin_remaining_tasks}. Best and second-best reported results are boldfaced and italicized. }
    \label{tab:robotwin_complementary_tasks}
    \setlength{\tabcolsep}{1pt}
    \renewcommand{\arraystretch}{1.22}
    \resizebox{1\textwidth}{!}{%
    \begin{tabular}{cccccccccccccc}
    \toprule
    \rowcolor{roboflblue}
    \color{white}{\textbf{Method}} &
    \color{white}{\taskname{adjust}{bot.}} & 
    \color{white}{\taskname{rank}{RGB}} & 
    \color{white}{\taskname{click}{alarm}} & 
    \color{white}{\taskname{click}{bell}} & 
    \color{white}{\taskname{dump}{bin}} & 
    \color{white}{\taskname{grab}{roller}} & 
    \color{white}{\taskname{hand}{mic}} & 
    \color{white}{\taskname{move}{pill}} & 
    \color{white}{\taskname{move}{card}} & 
    \color{white}{\taskname{open}{laptop}} & 
    \color{white}{\taskname{open}{micro.}} & 
    \color{white}{\taskname{bread}{bas.}} & 
    \color{white}{\taskname{burger}{fries}} \\ 
    \midrule
    \multicolumn{14}{c}{\texttt{CENTRALIZED TRAINING}} \\
    \midrule
    \rowcolor{roboflgray}
    \textsc{InternVLA} & \textit{99} & 91 & \textbf{85} & 91 & \textbf{98} & \textbf{100} & \textbf{95} & 87 & \textbf{100} & \textbf{98} & \textbf{91} & \textit{87} & \textbf{99} \\
    \textsc{Motus} & \textbf{100} & \textit{94} & 83 & 90 & 91 & \textbf{100} & 81 & \textit{91} & 97 & 92 & 80 & 81 & 93 \\
    \midrule
    \multicolumn{14}{c}{\texttt{FEDERATED LEARNING}\tiny\textsc{(LoRA/MoE)}} \\
    \midrule
    \rowcolor{roboflgray}
    \textsc{FedAvg} & \textit{99} & 91 & \textit{84} & \textit{93} & 95 & \textbf{100} & 89 & 83 & 98 & 90 & 84 & 82 & \textit{98} \\
    \textsc{FedMoE} & \textbf{100} & 91 & 76 & 91 & 94 & \textbf{100} & 82 & 80 & 95 & 88 & 65 & 80 & 97 \\
    \midrule
    \rowcolor{roboflgray}
    \textsc{ForgeVLA} & \textbf{100} & 90 & \textbf{85} & 92 & 94 & \textbf{100} & 82 & \textit{91} & \textit{98} & \textit{93} & 82 & 82 & 93 \\
    \textsc{FedVLA} & \textit{99} & 88 & 82 & 89 & 92 & \textit{99} & 83 & 83 & 96 & 90 & 78 & 82 & 96 \\
    \rowcolor{roboflgreen}
    \textbf{\robofl} & \textbf{100} & \textbf{95} & 83 & \textbf{96} & \textit{97} & \textbf{100} & \textit{92} & \textbf{92} & \textbf{100} & 90 & \textit{85} & \textbf{90} & 93 \\
    \midrule
    \rowcolor{roboflblue}
    \color{white}{\textbf{Method}} &
    \color{white}{\taskname{cans}{box}} & 
    \color{white}{\taskname{cont.}{plate}} & 
    \color{white}{\taskname{empty}{cup}} & 
    \color{white}{\taskname{place}{fan}} & 
    \color{white}{\taskname{obj.}{stand}} & 
    \color{white}{\taskname{phone}{stand}} & 
    \color{white}{\taskname{place}{shoe}} & 
    \color{white}{\taskname{shake}{bot.}} & 
    \color{white}{\taskname{shake}{horiz.}} & 
    \color{white}{\taskname{stack}{3 blo.}} & 
    \color{white}{\taskname{stack}{2 blo.}} & 
    \color{white}{\taskname{stack}{2 bow.}} & 
    \color{white}{\textbf{Overall}} \\
    \midrule
    \multicolumn{14}{c}{\texttt{CENTRALIZED TRAINING}} \\
    \midrule
    \rowcolor{roboflgray}
    \textsc{InternVLA} & \textbf{97} & 97 & \textbf{100} & \textbf{91} & 88 & \textit{94} & \textit{93} & \textit{99} & 97 & \textbf{87} & \textbf{100} & 98 & \textit{81.60} \\
    \textsc{Motus} & \textbf{97} & \textit{99} & \textbf{100} & 85 & 87 & 93 & 91 & \textbf{100} & \textbf{100} & 83 & \textit{99} & 98 & 80.96 \\
    \midrule
    \multicolumn{14}{c}{\texttt{FEDERATED LEARNING}\tiny\textsc{(LoRA/MoE)}} \\
    \midrule
    \rowcolor{roboflgray}
    \textsc{FedAvg} & \textit{94} & \textbf{100} & \textbf{100} & 86 & 84 & 91 & \textit{93} & \textbf{100} & 97 & \textit{84} & \textit{99} & \textbf{100} & 78.92 \\
    \textsc{FedMoE} & 85 & \textit{99} & \textbf{100} & 84 & \textit{89} & 91 & 86 & \textit{99} & \textit{99} & 80 & \textit{99} & \textbf{100} & 75.12 \\
    \midrule
    \rowcolor{roboflgray}
    \textsc{ForgeVLA} & 93 & \textbf{100} & \textbf{100} & 88 & 85 & \textit{94} & 92 & \textbf{100} & \textbf{100} & 81 & \textbf{100} & \textit{99} & 80.70 \\
    \textsc{FedVLA} & 89 & \textbf{100} & \textbf{100} & 83 & 84 & 90 & 90 & \textbf{100} & \textit{98} & 84 & 95 & \textbf{100} & 79.32 \\
    \rowcolor{roboflgreen}
    \textbf{\robofl} & \textit{94} & \textbf{100} & \textbf{100} & \textit{90} & \textbf{91} & \textbf{95} & \textbf{95} & \textbf{100} & \textbf{100} & 82 & \textbf{100} & \textbf{100} & \textbf{83.12} \\
    \bottomrule
    \end{tabular}
    }
\end{table*}

\subsection{RLBench}
\label{app:rlbench_protocol}

\paragraph{Task-silo manifest.} Table~\ref{tab:rlbench_task_silo_manifest} identifies the reported evaluation tasks, not client IDs. The training implementation provides task-disjoint \texttt{by\_task} partitions with four clients and a separate server.

\paragraph{Observation and action interface.} See Table~\ref{tab:benchmark_hyperparameters}. Unlike RoboTwin, each queued translation increment is added to the current end-effector position, while the predicted Euler angles specify an absolute world-frame orientation. The resulting pose is executed by a motion planner. Stored translation labels are already relative while preprocessing must not subtract the state again.

\paragraph{Sampling and update accounting.} The implementation uses the same shuffled-loader and update-counting conventions as RoboTwin. Table~\ref{tab:benchmark_hyperparameters} records launcher defaults.

\paragraph{Batch-size conventions.} The same per-client-update and per-server-rank conventions apply as in RoboTwin for the independent-client configuration. A grouped-client distributed run also requires the number of ranks per client to determine the effective local batch size.

\paragraph{Randomized initialization.} Simulator resets use the variation and seed in Table~\ref{tab:benchmark_hyperparameters}, without RoboTwin's expert-based candidate screening. Invalid actions and failures terminate the episode and count as failures. Expert actions and future observations are not supplied to the policy.

\begin{table}[t]
    \centering
    \caption{RLBench evaluation manifest. Rows identify evaluation tasks, not federated clients.}
    \vspace{0.5em}
    \label{tab:rlbench_task_silo_manifest}
    \setlength{\tabcolsep}{20pt}
    \resizebox{1\textwidth}{!}{%
    \renewcommand{\arraystretch}{1.15}
    \begin{tabular}{@{}lr>{\raggedright\arraybackslash}p{0.75\textwidth}@{}}
    \toprule
    Partition & Count & Tasks \\
    \midrule
    Client 0 & 1 & \texttt{take\_umbrella\_out\_of\_umbrella\_stand} \\
    Client 1 & 2 & \texttt{close\_fridge}, \ \texttt{toilet\_seat\_down} \\
    Client 2 & 2 & \texttt{close\_laptop\_lid}, \ \texttt{close\_box} \\
    Client 3 & 1 & \texttt{sweep\_to\_dustpan} \\
    \midrule
    Server & 2 & \texttt{put\_rubbish\_in\_bin}, \ \texttt{phone\_on\_base} \\
    \bottomrule
    \end{tabular}
    }
\end{table}

\section{Additional Experimental Details for Real-World Franka Tasks}
\label{app:franka_experimental_details}

\subsection{Implementation Details}
\paragraph{Task-silo manifest.} Table~\ref{tab:franka_task_silo_manifest} identifies the six reported single-arm Franka tasks and their client and server assignments. The implementation provides task-disjoint partitions among four clients and a separate server, with each client owning one task family and the server retaining the remaining two. Evaluation is performed via real-robot inference on the physical platform described in Appendix~\ref{app:franka_setup}.

\begin{table}[t]
    \centering
    \caption{Franka real-robot manifest. Rows identify evaluation tasks, not federated clients.}

    \vspace{0.5em}
    \label{tab:franka_task_silo_manifest}
    \setlength{\tabcolsep}{20pt}

    \resizebox{1\textwidth}{!}{%
    \renewcommand{\arraystretch}{1.15}
    \begin{tabular}{@{}lr>{\raggedright\arraybackslash}p{0.75\textwidth}@{}}
    \toprule
    Partition & Count & Tasks \\
    \midrule
    Client 0 & 1 & \texttt{stack\_cups} \\
    Client 1 & 1 & \texttt{screw\_bottle} \\
    Client 2 & 1 & \texttt{stamp\_seal} \\
    Client 3 & 1 & \texttt{test\_tube} \\
    \midrule
    Server & 2 & \texttt{adjust\_bottle}, \ \texttt{pour\_water} \\
    \bottomrule
    \end{tabular}
    }
\end{table}

\paragraph{Observation and action interface.} See Table~\ref{tab:benchmark_hyperparameters}. Unlike RoboTwin, each action is an absolute joint target executed directly by the robot, while queued targets are not added to the state at the start of the action queue. The policy consumes the external stand and wrist views, with the third view slot masked. State and action are both eight-dimensional, and auxiliary tactile and wrench channels are not supplied to the policy.

\paragraph{Sampling and update accounting.} The implementation uses the same shuffled-loader and update-counting conventions as RoboTwin and RLBench. Table~\ref{tab:benchmark_hyperparameters} records the schedule.

\paragraph{Batch-size conventions.} The same per-client-update and per-server-rank conventions apply as in RoboTwin and RLBench for the independent-client configuration. The client batch size is $32$ per local update, while the server batch size is $16$ per distributed rank.

\paragraph{Evaluation protocol.} Each method is evaluated in three rounds of 20 trials per task. Each real-robot trial is started and reset by the operator, which places the objects and fixtures in their nominal task configuration. The policy then executes in a closed loop on the physical robot, and a trial is scored as successful only if the task goal is reached; otherwise, the episode terminates and counts as a failure.

\subsection{Franka Single-Arm Platform}
\label{app:franka_setup}

\begin{figure}[t]
    \centering
    \includegraphics[width=0.7\textwidth]
        {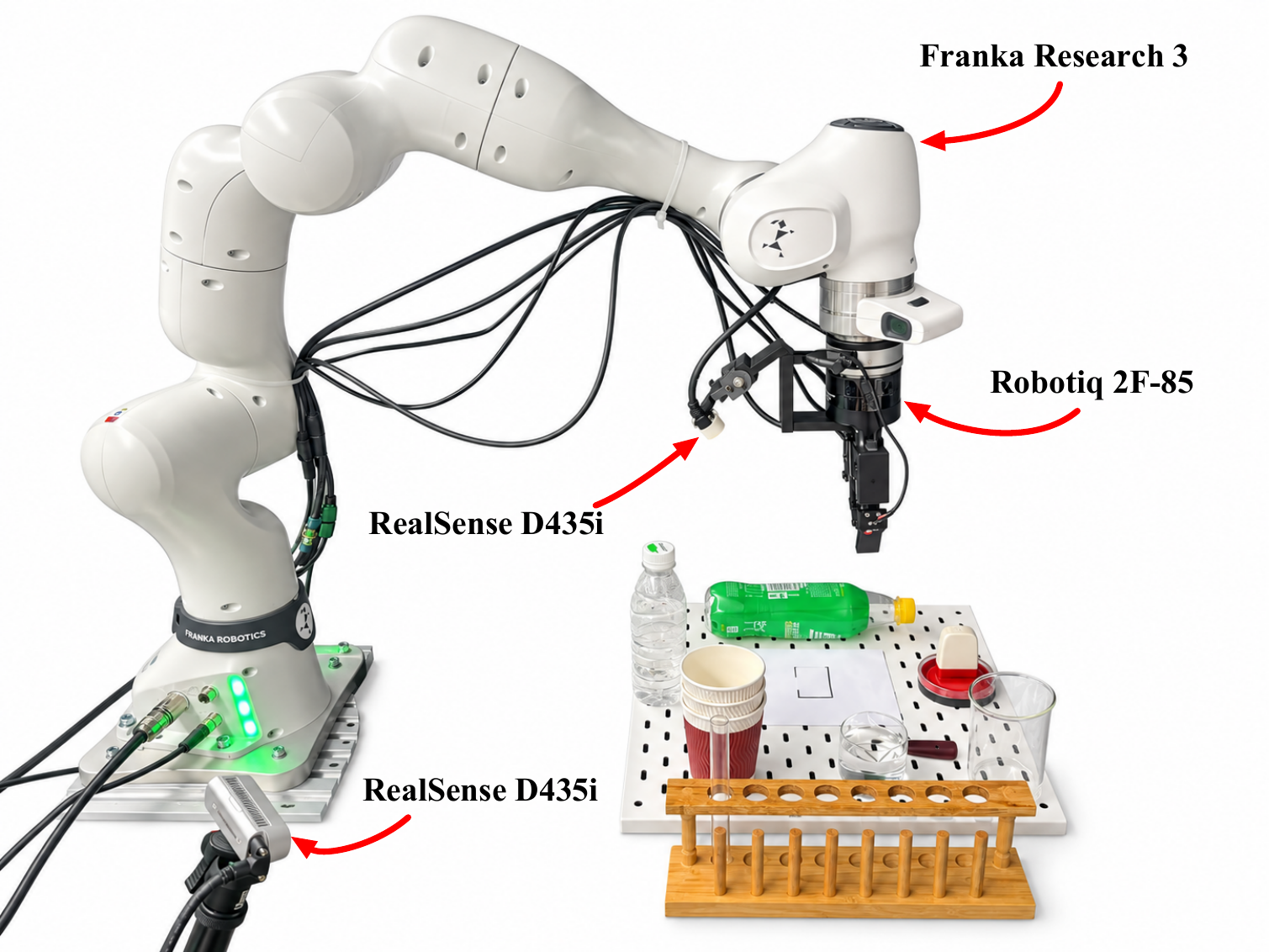}
    \caption{
        \textbf{Franka single-arm experimental setup.}
        The platform comprises a Franka Research~3, a Robotiq 2F-85 gripper, and two
        RealSense D435i cameras mounted on the wrist
        and an external stand, respectively.
        The tabletop contains the objects and fixtures
        used in the manipulation tasks.
    }
    \label{fig:franka_real_world_setup}
\end{figure}

\paragraph{Robot and gripper.}
Figure~\ref{fig:franka_real_world_setup} shows the hardware configuration for our Franka single-arm experiments. The platform consists of a Franka Research~3 manipulator
equipped with a Robotiq 2F-85 gripper. The robot is mounted beside a tabletop workspace containing the objects and fixtures used in the manipulation tasks.

\paragraph{Camera arrangement.}
Two RealSense D435i cameras provide complementary views of the workspace. One camera is mounted externally on a stand and faces the tabletop, providing an overview of the scene and object arrangement. The second camera is mounted on the robot wrist, providing a local view of the manipulation area as the arm moves. This configuration combines an external scene view with a close-range view of gripper-object interactions.

\paragraph{Workspace and task objects.}
The workspace contains a perforated board and
task-specific objects, including bottles, cups, a stamp and ink pad, glassware, and a test-tube rack. These objects support the six Franka tasks reported in the main text:
\textit{adjust bottle},
\textit{stamp seal},
\textit{stack cups},
\textit{screw bottle},
\textit{pour water}, and
\textit{test tube}.

\section{Additional Experimental Results}
\subsection{Complementary Quantitative Analysis
 for RoboTwin 2.0}
The complementary results explain the emphasis on challenging tasks in Table~\ref{tab:robottwin_remaining_tasks}. Most reported scores on this subset exceed 90\%, with many at or near 100\%, leaving limited headroom to distinguish methods. The main table therefore focuses on a lower-success subset where FedAvg scores below 80\%. This presentation makes \robofl's improvements on difficult manipulation tasks more visible, rather than relying on comparisons among near-saturated results, while the appendix preserves complete task coverage.

Despite the higher success rates and reduced room for improvement, \robofl \ remains highly competitive, achieving best or joint-best results on 16 of the 25 tasks and 100\% observed success on nine. Strong results across ranking, object placement, and bottle shaking show that its performance extends beyond the difficult cases highlighted in the main table. Together, the two subsets demonstrate that \robofl \ combines improvements on challenging tasks with strong performance on the more saturated portion of the benchmark.

In addition, Table~\ref{tab:mosaic_ablation} reports the overall success rate of \mosaic\ components on the full 50-task RoboTwin 2.0 benchmark. Installing task-trained adapters as expert branches with learned routing (vanilla \mosaic) reaches 81.94\%; adding routing distillation (\fard) raises this to 82.42\%; and adding consensus-weighted conversion (\pcea) reaches 83.12\%. The two external references put these numbers in context: federated averaging reaches 78.92\%, while a client-side-routed LoRA-MoE (FedMoE) reaches 75.12\%. Most of the gain, therefore, comes from retaining task-trained experts and learning their composition on the server (3.02\% over FedAvg), with \fard \ and \pcea \ contributing smaller but consistent additional improvements (0.48\% and 0.70\%, respectively). \fard \ and \pcea \ mainly affect the distribution of high-success tasks as shown in Section~\ref{sec:ablation}.

\begin{table}[t]
    \centering
    \caption{Ablation of \mosaic\ components on RoboTwin 2.0. All rows report overall success rates (\%) on all 50 tasks with 100 trials per task. The three \mosaic\ rows use the same federated pipeline and differ only in the server mechanism; FedAvg and FedMoE are external references.}
    \label{tab:mosaic_ablation}
    \setlength{\tabcolsep}{50pt}
    \renewcommand{\arraystretch}{1.15}
    \resizebox{1\textwidth}{!}{%
    \begin{tabular}{lc}
    \toprule
    \rowcolor{roboflblue}
    \color{white}{\textbf{Configuration}} & \color{white}{\textbf{Overall}} \\
    \midrule
    \rowcolor{roboflgray}
    FedAvg (averaged single adapter) & 78.92 \\
    FedMoE (client-side routed MoE) & 75.12 \\
    \midrule
    Vanilla \mosaic\ (task-trained experts + routing) & 81.94 \\
    \rowcolor{roboflgray}
    \quad $+$ \fard & 82.42 \\
    \quad $+$ \fard\ $+$ \pcea\ (\robofl) & \textbf{83.12} \\
    \bottomrule
    \end{tabular}
    }
\end{table}

\subsection{Quantitative Analysis for RLBench}
\begin{table*}[t]
    \centering
    \caption{RLBench task success rates (\%) on the 8-task single-view suite. Overall is the equally weighted mean across the eight tasks. Best and second-best results are boldfaced and italicized.}
    \label{tab:rlbench_single_view}
    \renewcommand{\taskname}[2]{{\ttfamily\makecell[c]{#1\\#2}}}
    \setlength{\tabcolsep}{4pt}
    \renewcommand{\arraystretch}{1.28}
    \resizebox{1\textwidth}{!}{%
    \begin{tabular}{cccccccccc}
    \toprule
    \rowcolor{roboflblue}
    \color{white}{\textbf{Method}} &
    \color{white}{\taskname{put}{rubbish}} & 
    \color{white}{\taskname{seat}{down}} & 
    \color{white}{\taskname{draw}{umbrella}} & 
    \color{white}{\taskname{close}{laptop}} & 
    \color{white}{\taskname{sweep}{dustpan}} & 
    \color{white}{\taskname{close}{fridge}} & 
    \color{white}{\taskname{close}{box}} & 
    \color{white}{\taskname{phone}{on base}} & 
    \color{white}{\textbf{Overall}} \\
    \midrule
    \multicolumn{10}{c}{\texttt{CENTRALIZED TRAINING}} \\
    \midrule
    \rowcolor{roboflgray}
    \textsc{InternVLA} & \textbf{60} & \textit{18} & \textit{34} & \textit{62} & \textit{66} & 46 & \textit{86} & \textbf{50} & \textbf{52.75} \\
    \textsc{Motus} & \textit{44} & \textit{22} & \textbf{36} & 58 & \textbf{72} & 58 & \textbf{92} & \textit{38} & \textit{52.50} \\
    \midrule
    \multicolumn{10}{c}{\texttt{FEDERATED LEARNING}\tiny\textsc{(LoRA/MoE)}} \\
    \midrule
    \rowcolor{roboflgray}
    \textsc{FedAvg} & 0 & 18 & 22 & 4 & 14 & \textbf{72} & \textbf{92} & 0 & 27.75 \\
    \textsc{FedMoE} & 0 & 12 & 16 & 14 & 22 & 64 & 70 & 8 & 25.75 \\
    \midrule
    \textsc{ForgeVLA} & 12 & 14 & 32 & 26 & 20 & \textit{70} & 84 & 16 & 34.25 \\
    \textsc{FedVLA} & 6 & 16 & 24 & 12 & 30 & 56 & 80 & 24 & 31.00 \\
    \rowcolor{roboflgreen}
    \textbf{\robofl} & 26 & \textbf{24} & 32 & \textbf{64} & 24 & 50 & 60 & \textbf{50} & 41.25 \\
    \bottomrule
    \end{tabular}
    }
\end{table*}

We further evaluate the methods in a lower-data, single-view RLBench setting with eight tasks and 50 trials per task. \robofl \ reaches $41.25\%$, the best among federated methods, improving over FL baselines. Unlike RoboTwin, where \robofl \ also surpasses centralized PEFT, it trails centralized InternVLA ($52.75\%$) and Motus ($52.50\%$) on RLBench. We hypothesize that the smaller single-view suite provides less task diversity and complementary evidence for expert composition, whereas centralized PEFT can directly pool all available examples. RoboTwin better represents the broad, task-heterogeneous distributed regime in which preserving specialized adapters is most valuable. These results therefore suggest that \robofl \ is particularly suited to large-scale distributed learning with complementary non-IID data.

\subsection{Qualitative Analysis}
\label{app:qualitative_analysis}

We visualize successful rollouts to examine whether the reported success rates correspond to sensible task behavior: approach and alignment, contact or grasp, and execution of the intended state change. Each task contributes four chronologically ordered keyframes from a single successful episode, and the frames are exported at their native resolution without cropping or enhancement. Because the evaluators record an observation before each executed action, the final frame may precede the release or contact that triggered the success flag, so visual inspection captures the executed behavior rather than confirming the terminal success event.

\begin{figure*}[t]
    \centering
    \includegraphics[width=\textwidth,height=0.86\textheight,keepaspectratio]{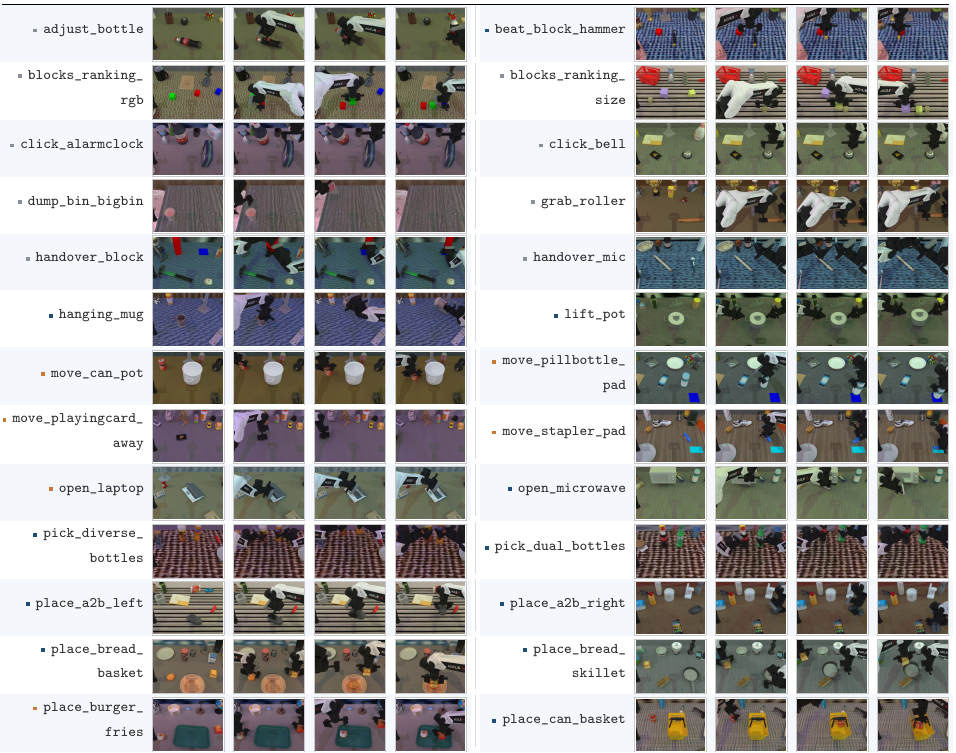}
    \caption{\textbf{Qualitative RoboTwin 2.0 rollouts (part 1 of 2).}
Each row shows two tasks. Every task is labeled with its name and contributes
four chronologically ordered keyframes from one successful rollout, illustrating
approach, contact or grasp, and execution of the intended state change. Frames
are exported at their native resolution without cropping or enhancement.}
    \label{fig:appendix_robotwin_qualitative}
\end{figure*}

\begin{figure*}[t]
    \centering
    \includegraphics[width=\textwidth,height=0.86\textheight,keepaspectratio]{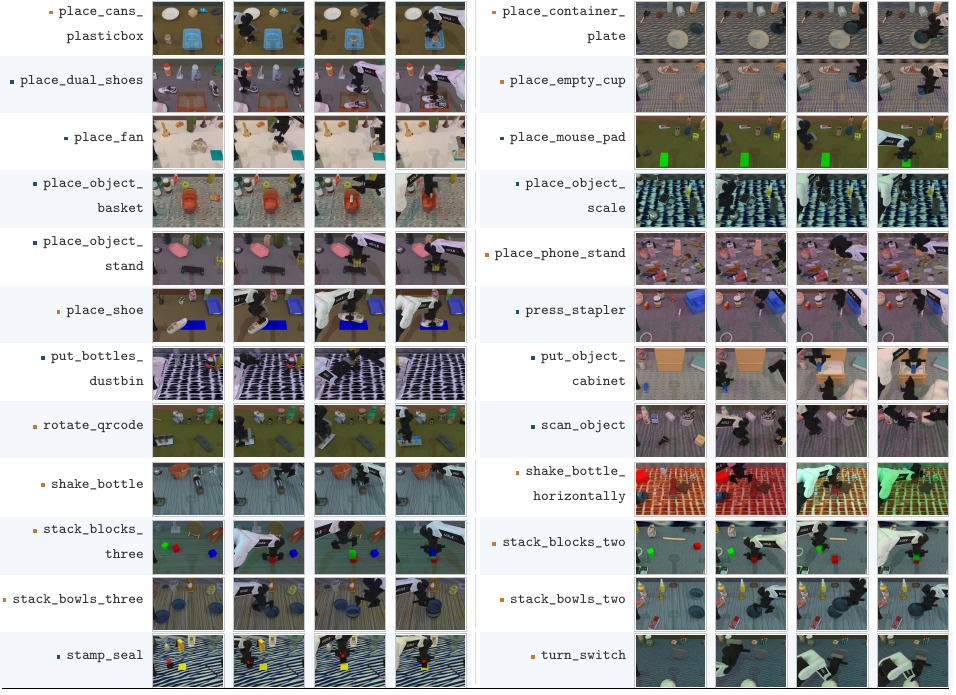}
    \caption{\textbf{Qualitative RoboTwin 2.0 rollouts (part 2 of 2).}
Each row shows two tasks. Every task is labeled with its name and contributes
four chronologically ordered keyframes from one successful rollout, illustrating
approach, contact or grasp, and execution of the intended state change. Frames
are exported at their native resolution without cropping or enhancement.}
    \label{fig:appendix_robotwin_qualitative_1}
\end{figure*}

\paragraph{RoboTwin 2.0.}
Figure~\ref{fig:appendix_robotwin_qualitative} and ~\ref{fig:appendix_robotwin_qualitative_1} show one successful rollout for each of the 50 RoboTwin 2.0 tasks. Across these rollouts, the policy exhibits a consistent three-phase structure: an alignment phase in which the end effector approaches the target, a contact or grasp phase, and a transport or placement phase. Articulated and contact-rich tasks such as \textit{open microwave}, \textit{close laptop lid}, and \textit{rotate qrcode} show the gripper establishing contact and then applying a sustained motion rather than a ballistic strike. Multi-object tasks such as the stacking and ranking families, \textit{place bread basket}, \textit{place cans plasticbox}, and \textit{place dual shoes} involve sequential subgoals, in which the second object is manipulated only after the first has been positioned. The tasks that remain difficult in the quantitative results, notably \textit{hanging mug} and \textit{lift pot}, are also visually harder: they require precise two-sided or handle-based contact under substantial occlusion, and their successful rollouts show a slower, more tentative approach. The remaining failures in the report are therefore concentrated in fine-grained contact and long-horizon placement rather than in gross reaching or scene understanding.

\begin{figure*}[t]
    \centering
    \includegraphics[width=\textwidth]{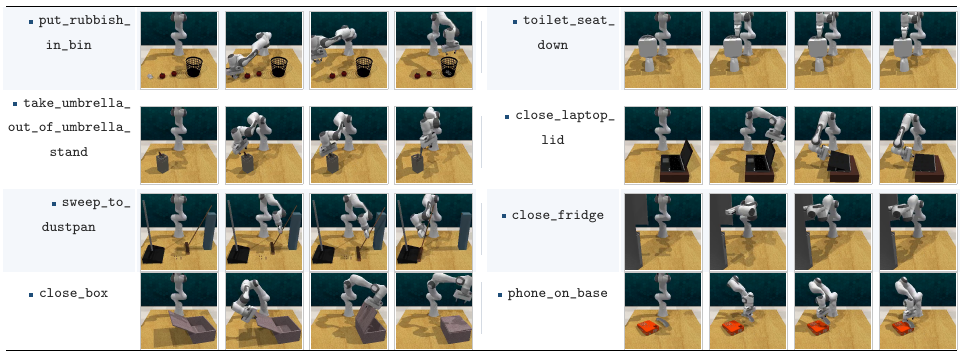}
    \caption{\textbf{Qualitative RLBench rollouts.}
Each row shows two tasks, each labeled with its name and the number of successful
episodes out of 50, and includes up to four chronologically ordered keyframes from
one successful rollout. All rollouts use the single-view interface with the front
camera at variation~0 and the eight-action chunk-queue protocol. The displayed
episode for each task was selected for visual clarity rather than being the longest
or most representative rollout.}
\vspace{-2em}
    \label{fig:appendix_rlbench_qualitative}
\end{figure*}

\paragraph{RLBench.}
Figure~\ref{fig:appendix_rlbench_qualitative} shows successful rollouts for the eight-task single-view suite under the evaluation protocol of Table~\ref{tab:benchmark_hyperparameters}: front camera, variation~0, and an eight-action chunk queue. 
The rollouts reveal several recurring behaviors. Closing tasks, including \textit{close Box}, \textit{close fridge}, and \textit{close laptop lid}, are executed as a short contact-and-push sequence: the gripper reaches the lid or door, establishes contact, and then drives it through its range in a single continuous motion. Grasping-and-relocation tasks such as \textit{put rubbish in bin} and \textit{take umbrella out of umbrella stand} show a descending approach and a closed grasp before the object is lifted clear. Tool-use and placement tasks, \textit{sweep to dustpan} and \textit{phone on base}, require the gripper to align with a narrow target region before the state change.

\subsection{Failure Analysis}
\label{app:failure_analysis}

\begin{figure*}[t]
    \centering
    \includegraphics[width=\textwidth]{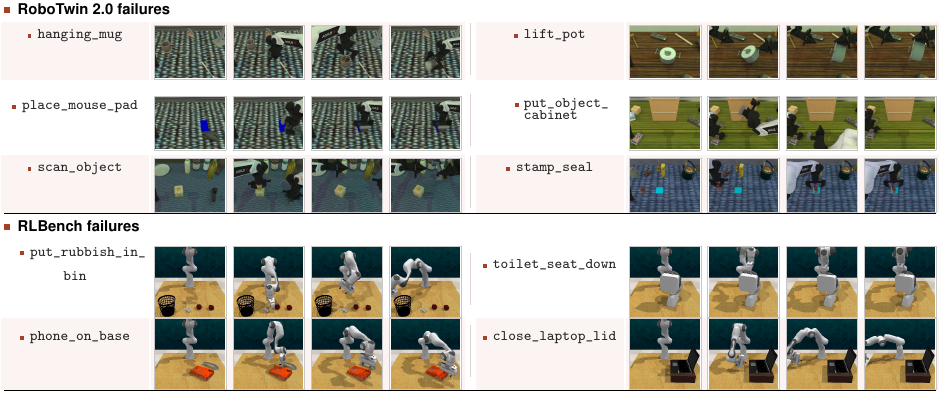}
    \caption{\textbf{Failure cases across benchmarks.}
    RoboTwin 2.0 failures (top) and RLBench failures (bottom) contribute four chronologically ordered keyframes from one failed rollout: initial scene, first meaningful attempt, visible deviation, and final failing state. Because observations are recorded before actions and the video ends at the failure, the last frame shows the episode-terminated state.}
    \label{fig:appendix_failure}
\end{figure*}

\begin{figure*}[t]
    \centering
    \includegraphics[width=\textwidth]{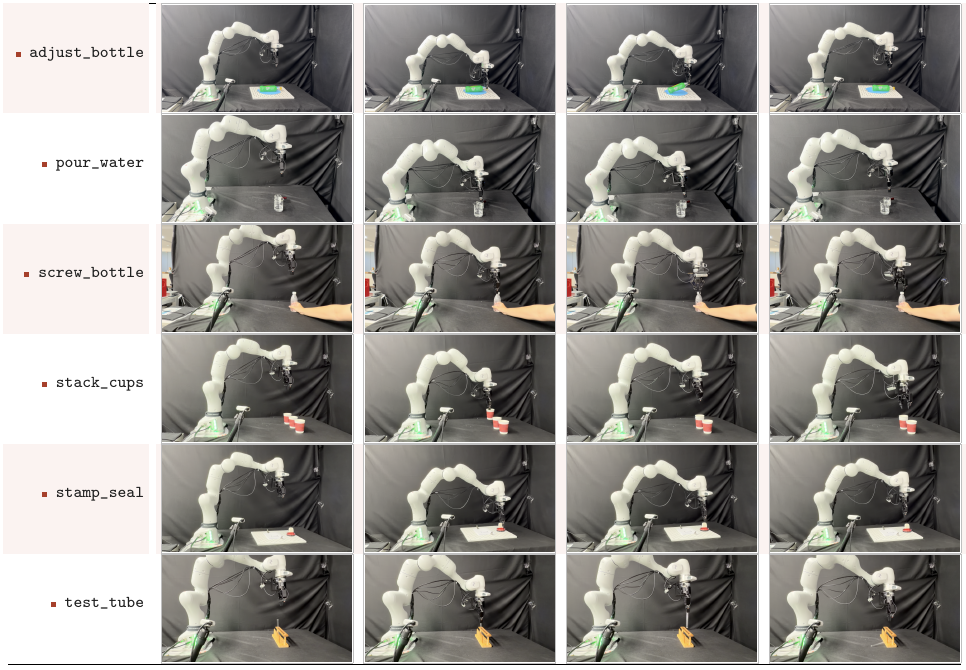}
    \caption{\textbf{Franka failure cases.}
    Failed rollouts for six Franka tasks, with each row contributing four chronologically ordered keyframes (initial scene, first attempt, visible deviation, final state).}
    \label{fig:appendix_franka_fail}
\end{figure*}

Despite the overall gains, we observe recurring failures in both simulation and hardware, as illustrated in Figures~\ref {fig:appendix_failure} and~\ref {fig:appendix_franka_fail}. \textbf{1) Incomplete Object Engagement:} In \textit{lift pot} and \textit{hanging mug} the gripper topples or tilts the object instead of lifting or hanging it, and on Franka \textit{screw bottle} never achieves threaded engagement. This suggests that a skill absent from every client family must be synthesized by expert composition, which can fall back to a blend of adjacent experts. \textbf{2) Stalled or Abandoned Motion:} \textit{stamp seal} hovers without pressing, \textit{scan object} drifts away after approaching, and on Franka \textit{pour water} and \textit{test tube} stop short of the manipulation, consistent with weak understanding-generation consensus under self-occlusion leaving the router under-constrained. \textbf{3) Misplaced End State:} \textit{put object cabinet} drags the object instead of stowing it, and \textit{stack cups} lands beside the base, indicating that rank-limited conversion attenuates the expert-specific component of a terminal motion. The RLBench failures in Figure~\ref{fig:appendix_failure} (bottom) differ, all ending in a motion-planning error (\emph{InvalidActionError}) rather than a policy error. Overall, these cases highlight that skills absent from all clients, ambiguous routing, and lossy conversion remain the main bottlenecks, which we plan to instrument in future work.

\section{Statistical Reporting Protocol}
\label{app:statistics}

Recent VLA and WAM work typically reports success rates over a fixed number of execution trials from a single training run rather than averaging over multiple training seeds on the simulation: for example, Motus evaluates each RoboTwin 2.0 task over $100$ execution trials from one fine-tuned model~\citep{bi2026motus}, and comparable single-run protocols are used for InternVLA-A1~\citep{cai2026internvla} and ForgeVLA~\citep{zhou2026forgevla}. We follow this convention: all simulation results are obtained from a single fixed training seed and a single evaluation seed, with $100$ (RoboTwin 2.0) and $50$ (RLBench) trials per task, and the real-robot protocol is a fixed, scripted sequence. This keeps our results directly comparable to the benchmarks and baselines that we report. Multi-seed training with confidence intervals would strengthen the evidence for these smaller effects and is a natural direction for future work.

\section{Limitations}
Our study has several limitations. First, all experiments use a single backbone (InternVLA-A1-3B) and one aligned MoT architecture. Second, we target task heterogeneity: clients differ in task distributions but share the embodiment, sensor suite, and action space, so cross-embodiment and cross-modal heterogeneity remain open questions. Finally, simulation results use a single evaluation seed, and the real-robot results use a fixed scripted protocol with 20 trials per task. 

\end{document}